\documentclass[%
 reprint,
 amsmath,amssymb,
 aps,
 prx,
]{revtex4-2}

\usepackage{tikz}
\usepackage{natbib}
\usepackage{graphicx}
\usepackage{dcolumn}
\usepackage{bm}
\usepackage{hyperref}

\newcommand\defeq{\mathrel{\overset{\makebox[0pt]{\mbox{\normalfont\tiny\sffamily def}}}{=}}}
\def\E{\mathbb{E}}
\def\D{\mathcal{D}}

\begin{document}

\title{Learning curves for deep neural networks:\\ A field theory perspective}

\author{Omry Cohen}
 \email{omry.cohen@mail.huji.ac.il}
\author{Or Malka}%
 \email{or.malka@mail.huji.ac.il}
\author{Zohar Ringel}%
 \email{zohar.ringel@mail.huji.ac.il}
\affiliation{The Racah Institute of Physics, The Hebrew University of Jerusalem.}

\date{\today}

\begin{abstract}
In the past decade, deep neural networks (DNNs) came to the fore as the leading machine learning algorithms for a variety of tasks. Their raise was founded on market needs and engineering craftsmanship, the latter based more on trial and error than on theory. While still far behind the application forefront, the theoretical study of DNNs has recently made important advancements in analyzing the highly over-parameterized regime where some exact results have been obtained. Leveraging these ideas and adopting a more physics-like approach, here we construct a versatile field-theory formalism for supervised deep learning, involving renormalization group, Feynman diagrams and replicas. In particular we show that our approach leads to highly accurate predictions of learning curves of truly deep DNNs trained on polynomial regression problems. It also explains in a concrete manner why DNNs generalize well despite being highly over-parameterized, this due to an entropic bias to simple functions which, for the case of fully-connected DNNs with data sampled on the hypersphere, are low order polynomials in the input vector. Being a complex interacting system of artificial neurons, we believe that such tools and methodologies borrowed from condensed matter physics would prove essential for obtaining an accurate quantitative understanding of deep learning.  
\end{abstract}

\maketitle


\section{Introduction}
\label{Introduction}

Deep artificial neural networks (DNNs) have been rapidly advancing the state-of-the-art in machine learning, showing human and sometimes super-human performance in image recognition \cite{ImageNet2020}, speech recognition \cite{Saon2017}, reinforcement learning \cite{DeepMindChess} and natural language processing tasks \cite{Kalchbrenner2016}. Their raise to prominence was largely results-driven, with little theoretical support or guarantee \cite{OpeningTheBlackBoxUsingInformation}. Such mode of invention is very different from how, say, the transistor was discovered, and more akin to how new materials, such as lithium-ion batteries, are discovered. Indeed, being huge interacting systems of artificial neurons, DNNs are more analogous to a complex meta-material than to an electronic component \cite{Hexner2019}. Due to this complexity, a general theory of deep learning with predictive power is still lacking.


Notwithstanding, recently several results were obtained in the highly over-parameterized regime \cite{Daniely2016, Jacot2018} where the role played by any specific DNN weight is small. This facilitated the proofs of various bounds \cite{AllenZhu2018, CaoB2019, Cao2019} on generalization for shallow networks and, more relevant for this work, two correspondences between fully-trained DNNs and a different type of inference models called Gaussian Processes (GPs) \cite{Rasmussen2005}. As shown below, these can be thought of as non-interacting scalar field-theories with disorder and a non-local action. 

The first such correspondence \cite{Jacot2018} between GPs and trained DNNs is known as the Neural Tangent Kernel result, which we would refer to here as the NTK correspondence. It holds when highly over-parameterized DNNs are initialized according to standard practice and trained with Mean-Square-Error (MSE) loss at vanishing learning rate and without weight decay. 

The second correspondence \cite{Gadi2020} (the NNSP, Neural Networks Stochastic Process correspondence) applies when DNNs are trained using a similar protocol which involves random noise, roughly mimicking the Stochastic Gradient Descent (SGD) optimization, as well as weight-decay. It relates the outputs of the trained DNN to a Stochastic Process (SP) which, in the highly over-parameterized limit, tends to a GP. It thus yields an additional training protocol, complementary in some ways to the previous one, which is analytically tractable. 

How much of deep learning can be explained through such correspondences remains to be seen. On the one hand, some aspects such as learning sharp filters (features) in the first DNN layers, seem out of reach as specific DNN weights change only infinitesimally in the NTK case and remain largely random, apart from a small bias, in the NNSP case. In addition, learning in the NTK regime, sometimes dubbed "lazy-learning", often lags behind state-of-the-art training protocols (see \onlinecite{Ghorbani2020} and Refs. therein) where finite learning-rates, widths, and mini-batches are used. On the other hand, lazy-learning or more generally GP methods are being extended and improved \cite{Arora2019Harnessing, LiZhiyuanWang2019, Lee2018,Lee2020width} by importing technologies such as pooling and data-augmentation. Currently GP models corresponding to DNNs are competitive with deep learning on the UCI datasets \cite{Arora2019Harnessing} as well as Fashion-MNIST \cite{LiZhiyuanWang2019}, whereas on the CIFAR-10 dataset the performance of the best GPs currently lags $5\%$ behind celebrated DNNs such as AlexNet \cite{LiZhiyuanWang2019,Lee2020width}, while surpassing pre-AlexNet non-deep methods by $8\%$ \cite{CIFARSOTA2010}. In addition, there is the prospect of extending these correspondences by include non-linearities coming from finite-width \cite{Gadi2020} and finite-learning rates \cite{GuyGur2020,WARMING1974159} settings. These results and prospects invite further study of how such DNNs trained in the NTK and NNSP regimes make predictions.

In this work we introduce a versatile field-theory formalism for analyzing deep neural networks, which involves replicas, Feynman diagrams, and renormalization group techniques. In its most basic version, studied in depth below, it applies to DNNs trained using the protocols for which the NTK and NNSP correspondence hold exactly and lead to GP models. For these cases we provide expressions for the generalization power of fully-connected DNNs in the form of learning curves. These learning curves depend on the dataset distribution and the target which we learn. For uniform datasets on the hyper-sphere and any target function, our learning curves become fully explicit and provide a clear picture of how such DNNs generalize. This includes the more challenging case of the NTK correspondence where certain infinities in the action are removed by our renormalization group transformation. To the best of our knowledge, the accuracy at which our learning-curves capture the empirical ones far exceeds the current theoretical state of the art. 

In addition our formalism can also accommodate various extension of these correspondences.  For the case of the NNSP correspondence, we can work with loss functions different than MSE as well as corrections to the infinite over-parameterization limit. Furthermore, recent results on extensions of the NTK correspondence \cite{GuyGur2020}, suggest that high-learning-rate leads to a renormalized NTK correspondence whose performance can again be analyzed using our approach. Such extensions may prove useful in addressing the gap \cite{Arora2019OnExact}, between GPs and their DNN counterparts.    

We hope that the results and formalism introduced here would aid in developing a more physics-like paradigm for studying DNNs, complementary to the proof-based approach common in theoretical computer science (see also \cite{Zdeborov2020, Levine2019, Li2018Exploring, Becker2020, Tramel2018}). Such a paradigm should fill in the gap, typically large in complex systems, between what can be predicted following some reasonable assumptions and what can be proven rigorously.

This paper is structured as followed. In section \ref{Sec:Background} we provide the necessary background on Deep neural networks, Gaussian processes and the correspondences between the two. Section \ref{Sec:FieldTheoryFormulation} describes our novel field theory approach and analytical results. Section \ref{Sec:Uniform} considers the case of uniformly distributed data on the hypersphere, where further analytical simplifications can be carried. Section \ref{Sec:RG} introduces the RG approach used to tackle the noiseless NTK case. Section \ref{Sec:ConcreteExample} applies our results to concrete examples and compares them with empirical results. Section \ref{Sec:Hyper-parameterOptimization} shows how our results can be used to perform efficient hyper-parameters optimization on actual DNNs, and Section \ref{Sec:Discussion} summarizes the results and discusses possible directions for future work.

\section{Prior works}
\label{Sec:PriorWorks}

Learning curves for GPs have been analyzed using a variety of techniques (see \citep{Rasmussen2005} for a review) most of which focus on a GP-teacher averaged case where the target/teacher is drawn from the same GP used for inference (matched priors) and is furthermore averaged over. Fixed-teacher or fixed-target learning curves have been analyzed using a grand-canonical/Poisson-averaged approach \citep{Malzahn2001} similar to the one we used. However, their treatment of the resulting partition function was variational whereas we take a different, perturbation-theory based, approach. In addition, previous cited results for MSE-loss break in the noiseless limit \citep{Malzahn2001}. To the best of our knowledge, noiseless GPs learning-curves have been analyzed analytically only in the teacher-averaged case and limited to the following settings: For matched priors, exact results are known for one dimensional data \citep{Williams2000, Rasmussen2005} and two dimensional data with some limitations of how one samples the inputs (in the context of optimal design) \citep{ritter2007average, ritter1996}. In addition \citep{Micchelli1979} derived a lower bound on generalization. For noiseless inference with partially mismatched-priors (matching features, mismatching eigenvalues) and at large input dimension the teacher and dataset averaging involved in obtaining learning curves, has been performed analytically and the resulting matrix traces analyzed numerically \citep{Sollich2001}. Notably none of these cited results apply in any straightforward manner in the NTK-regime.

Considering kernel eigenvalues, explicit expression for the features and eigenvalues of dot-product kernels ($K=K(x \cdot x')$) were given in \citep{dotProdKer}. The fact that the $l$-th eigenvalue of such kernels scales as $d^{-l}$ ($d$ being the input dimension), which we used in our derivation of the bound, has been noticed in \citep{Sollich2001}. Kernels with a trimmed spectrum where the spectrum is trimmed after the first $r$'s leading eigenvalues, has previously been suggested as a way of reducing the computational cost of GP inference \citep{FerrariTrecate1998}. In contrast we trim the Taylor expansion of the kernel function rather than the spectrum (which has a very different effect) and show that an effective observation noise compensates for our trimming/renormalization procedure.

Several interesting recent works give bounds on generalization \citep{AllenZhu2018, CaoB2019, Cao2019} which show $O(1/\sqrt{N})$ asymptotic decay of the learning-curve (at best). In contrast our predictions are typically well below this bound.

\section{Theoretical background}
\label{Sec:Background}

\subsection{DNNs, expected error, and learning curves}
\label{Subsec:DNNs}

We begin with the standard definitions of DNNs as they apply to this work. While the majority of this work is applicable to many network architectures, we will focus on a simple feed forward network for the sake of simplicity. A fully connected feed forward DNN with $L$ hidden layers of width $n_l$ for $l=1,\ldots,L$ and readout layer $n_{L+1}=k$ is a function $f$ defined recursively by:

\begin{align}
    \left\{ \begin{array}{l}
h^{l+1}=x^{l}W^{l+1}+b^{l+1}\\
x^{l+1}=\phi\left(h^{l+1}\right)\\
f\left(x;W^{1},\ldots,W^{l},b^{1},\ldots,b^{l}\right)=x^{L+1}
\end{array}\right.
\end{align}

where $\phi$ is a point-wise activation function, $x^0 \in \mathbb{R}^{d}$ is the input of the network and $W^{l+1}\in\mathbb{R}^{n_{l}\times n_{l+1}}$, $b^{l+1}\in\mathbb{R}^{n_{l+1}}$ are trainable weights and biases, which will be collectively referred to as weights from here on. Each component of the weights is usually initialized randomly from a normal distribution ${\cal N}\left(0,\sigma_{w}^{2}\right)$ for the weights and ${\cal N}\left(0,\sigma_{b}^{2}\right)$ for the biases. 

In the usual setting one starts with a training set -- a set of input points $D=\left\{x_n\right\}^{N}_{n=1}$ where $x_n \in \mathbb{R}^{d}$ along with their labels $\left\{l_n\right\}^{N}_{n=1}$ where $l_n\in\mathbb{R}^k$. One then picks weights for the network by minimizing a loss function ${\cal L}\left(f\left(D\right),\left\{ l_{n}\right\} \right)$ which compares the values of network function over $D$ to the labels $\left\{l_n\right\}^{N}_{n=1}$, assigning a smaller value to points where the network function and labels are similar. One then finds weights which minimize the loss by some variation of gradient descent , usually stochastic gradient descent (SGD) wherein one approximates the gradient at each iteration using a random batch of the training set (see \cite{nielsen2015neural} for details). The performance of the network is then evaluated by computing the loss function over a set of labeled points, different from the training set, known as the test set. This is known as the test error of the network, and is used as a proxy for the \emph{expected error} -- the loss averaged over draws from the dataset distribution.

One of the most detailed objects quantifying the performance of a machine learning algorithm, and the main focus of this work, is its learning-curve -- a graph of how the expected error diminishes with the number of data points ($N$). There are currently no analytical predictions or bounds we are aware of for DNN learning-curves which are tight even just in terms of their scaling with $N$, let alone tight in an absolute sense (see App. \ref{Sec:PriorWorks})




\subsection{Gaussian processes regression}
\label{Subsec:GPR}

In this work we will investigate the properties of DNNs by their correspondence with GPs. We supply here some standard definitions of GPs and their usage in regression tasks. Regression here simply means approximating a function ($g(x)$) based on discrete samples ($\{ g(x_n) \}_{n=1}^{N}$). A GP is commonly defined as a stochastic process of which any finite subset of random variables follow a multivariate normal distribution \cite{Rasmussen2005}. In a similar fashion to multivariate normal variables, GPs are also determined by their first and second moments. The first is typically taken to be zero, and second is known as the covariance function or the kernel $K_{xx'} = \E\left[f(x) f(x')\right]$, where $\E[\cdot]$ here denotes expectation with respect to the GP distribution. The main appeal of GPs is that Bayesian Inference with GP priors is tractable \cite{Rasmussen2005}. In GP inference we use the mean of the GP distribution conditioned on the data (posterior) as the predictor $g^*$, and it is given by:

\begin{align}
\label{Eq:GPPred}
g^*(x_*) &= \sum_{n,m=1}^N K_{x_*,x_n} [ K(D) + \sigma^2 I]^{-1}_{nm} l_m  
\end{align}

where $x_*$ is a new data point, $l_m$ are the training targets, $x_n$ are the training data-points, $[K(D)]_{nm} = K_{x_n,x_m}$ is the covariance-matrix (the covariance-function projected on the training dataset $D$), and $\sigma^2$  is the variance of the assumed Gaussian noise of the labels, which also acts as a regulator of the prediction. Some intuition for this formula can be gained by verifying that in the noiseless case ($\sigma^2=0$) the prediction at some training point $x_* = x_q$ coincides with that point's label $g^* = l_q$. 

The quantity of interest in this paper, which we define now, is the expected error averaged over all the possible datasets. Throughout this paper we will assume that both train and test points are drawn from a probability measure $d\mu_{x}=P(x)dx$. With this in mind, we define the expected error of a prediction $g^*$ as

\begin{equation}
\label{Eq:predGen}
    \left\Vert g-g^*\right\Vert^2 = \int d\mu_x \left(g(x)-g^*(x)\right)^2
\end{equation}

Note that $g^*$ is itself a function of $N$ draws from $\mu$ which make up the training set $D_N$. Our quantity of interest, the dataset averaged expected error (DAEE), is Eq. \eqref{Eq:predGen} averaged over the ensemble of all possible $N$ sized training sets. We denote this average as $\langle \cdot \rangle_{D_N}$, so the DAEE is given by $\langle \left\Vert g^* - g \right\Vert^2 \rangle_{D_N}$. The learning curve is the dependence of the DAEE on $N$. We see that in order to calculate learning curves, one needs to calculate quantities like $\langle g^* \rangle_{D_N}$ and $\langle g^{*2} \rangle_{D_N}$.

Equation \eqref{Eq:GPPred} determines the predictions, and therefore the learning-curves, but it is not very convenient for analytic exploration of the expected predictions. This fact is due to the (potentially very) large matrix inversion involved, and the additional averaging over $D_N$ required. Nonetheless, there are some approximations for the expected prediction $\langle g^*\rangle_{D_N}$. The most famous of which is the equivalence kernel (EK) result \cite{Rasmussen2005}:

\begin{align}
\label{Eq:EK}
\langle g^*(x)\rangle_{D_N} \approx g^{*}_{EK,N}(x) = \sum_n \frac{\lambda_n}{\lambda_n + \frac{\sigma^2}{N}} g_n\phi_n(x)
\end{align}

Where $\lambda_n$ and $\phi_n(x)$ here are the eigenvalues and eigenfunctions of the kernel w.r.t the input probability measure $\mu$, and $g(x)=\sum_n g_n \phi_n(x)$ is the target function. One notices immediately that this approximation breaks down completely in the noiseless case where Eq. \eqref{Eq:EK} implies perfect estimation of the target with just one data point. To gain some intuition as to why having $\sigma^2=0$ hinders predictions of $\langle g^*\rangle_{D_N}$ one can view it as a hard constraint ($f(x_n)=g(x_n)$), and hard constraints are typically less tractable than soft ones. In a related view, finite $\sigma^2$ can be seen as a form of averaging which smooths and regulates analytical expressions making them more tractable. Another limitation of the EK result is that (to the best of our knowledge) there is no systematic way to extend it in orders of $1/N$ and get a more detailed picture of generalization in GP regression (GPR).

\subsection{From DNNs to GPs through Langevin dynamics} 
\label{Subsec:Langevin}
Here we review, for completeness, several recent correspondences between DNNs and GPs. It has long been known \cite{neal2012bayesian} that randomly initialized, infinitely wide DNNs with i.i.d weights are equivalent to samples from a GP known as the neural network GP (NNGP). More recently it was shown that training only the last layer of a network with gradient decent is equivalent to posterior sampling of the NNGP \cite{Lee2019}, and consequently averaging the prediction of many networks trained on the same dataset is equivalent to GPR. Turning to more standard training of the entire DNN, it has been recently  established \cite{Jacot2018} that fully training a network with vanishing learning rate for infinitely long time and MSE loss yields the same predictions as a noiseless GPR with a different kernel, the neural tangent kernel (NTK), along with an additional initialization dependent term. Averaging over many initialization seeds gives an exact correspondence with a GP whose kernel is the NTK.   

Recently, another novel correspondence between DNNs and GPs has been introduced \cite{Gadi2020}. Due to its simplicity we shall re-derive it here. Consider the training of a DNN using gradient descent (full-batch SGD) with weight decay and added white noise, in the limit of vanishing learning rate. For sufficiently small learning rate, and making the reasonable assumption that the gradients of the loss are globally Lifshitz, the SGD equations are ergodic and converge to the same invariant measure (equilibrium distribution) as the following Langevin equation \cite{MATTINGLY2002, risken1996fokker}
\begin{equation}
\label{Eq:Langevin}
\frac{d w_{i}}{d t}= -\partial_{w_{i}}\left(\mathcal{L}[z_{W}]+\sum_{j} \frac{T w_{j}^{2}}{2 \sigma_{w}^{2}} \right)+\sqrt{2 T} \xi_{i}(t)
\end{equation}
where $\xi_{i}(t)$ being a set of Gaussian white noise ($\left\langle\xi_{i}(t) \xi_{j}\left(t^{\prime}\right)\right\rangle=\delta_{i j} \delta\left(t-t^{\prime}\right)$), $T$ accounting for the strength of the noise, $w_i$ being the set of networks parameters ($W$), $z_{W}$ is the network output for a given configuration of $W$ and $\mathcal{L}$ is the loss function. The equilibrium-distribution or invariant-measure describing the steady state of the above equation is the Boltzmann distribution \cite{risken1996fokker} $P(W) \propto e^{-\frac{1}{2\sigma^2}\mathcal{L}\left[z_{W}\right] -\frac{1}{2\sigma_w^2} \sum_{i} w_{i}^{2} }$ with $T=2\sigma^2$. Notably, various works argue that at low learning rates, discrete SGD dynamics running for long enough times, reaches the above equilibrium, approximately \cite{Mandt2017, Welling2011}.

Next, we adopt the approach of \cite{Jacot2018} and describe the dynamics in function space ($f$) instead of weight space ($W$). Using the Boltzmann distribution described above, the post-training probability density function for some function $f$ is given by
\begin{align}
\label{Eq:Langevin_PDF}
    P[f]&=\int dW P(W) \delta[f-z_W] \\ \nonumber
    &\propto e^{-\frac{1}{2\sigma^2}\mathcal{L}\left[f\right] } \int dW e^{-\frac{1}{2\sigma_w^2} \sum_{i} w_{i}^{2} } \delta[f-z_W] \\ \nonumber
    &\propto P_{nd}[f] e^{-\frac{1}{2\sigma^2}\mathcal{L}\left[f\right] }
\end{align}
where we identify $P_{nd}[f]\propto \int dW e^{-\frac{1}{2\sigma_w^2} \sum_{i} w_{i}^{2} } \delta[f-z_W]$ as the distribution of the output of the network after being trained with no data (or equivalently, a vanishing loss function). $\delta[...]$ is a {\it functional} delta function, which can be thought of as the limit of a large product of regular delta-functions on each Fourier component of the argument. As we will discuss, for an infinitely over-parameterized network $P_{nd}$ coincides with the prior of a NNGP with the weights and biases variance determined by training parameters rather than by initialization. However, for finite over-parameterization it becomes a more generic stochastic process determined by the neural network (an NNSP). In \citep{Gadi2020}, the leading finite-width corrections were calculated and shown to result in $f^4$ corrections to the prior.  

Clearly the practical use of the above result hinges on how quickly the dynamics mixes or reaches ergodicity. While ergodicity in its full sense (for any weight-space observable) seems unrealistic in this non-convex scenario, reaching ergodicity in the mean of the outputs of the DNNs (for low order polynomials in $f(x)$) may be quicker. This milder form of ergodicity was shown numerically for fully-connected DNNs trained on regression problems similar to those studied here as well as CNNs trained on CIFAR-10 \cite{Gadi2020}.  

From now on we shall focus on the infinite over-parameterized limit.  

\section{Field theory formulation of GP learning-curves} 
\label{Sec:FieldTheoryFormulation}

\subsection{Rephrasing GPs as a field theory}
We begin by phrasing inference with GPs in the language of field theory. To this end we first write a Gaussian distribution over the space of functions that leads to a two point correlation function equal to $K_{xx'}$. This is given by 
\begin{gather}
    \label{eq:PDF_nodata_propto}
     P_0\left[f\right]\propto e^{-\frac{1}{2}\left\Vert f\right\Vert _{K}^{2}} \\ \nonumber
     \left\Vert f\right\Vert _{K}^{2}=\int d \mu_x d \mu_{x'} f(x) K^{-1}(x,x') f(x')
\end{gather}
where $K^{-1}(x,x')$ is the inverse kernel function, meaning that $\int d \mu_{x'} K(x,x') K^{-1}(x',x'')=\delta(x-x'')/P(x)$ where $d\mu=P(x)dx$. This formalism is sometimes referred to as Information Field Theory (IFT) \cite{Ensslin:1111177}.

A different viewpoint on $P_0[f]$ comes from viewing $f(x)$ as the outputs of a wide DNN with weights drawn from an i.i.d Gaussian distribution $P_0(W)$. It is well known \cite{Saul2009} that correlations between the outputs of random DNNs are Gaussian and governed by some kernel, $K_{xx'}$. This kernel is determined, in a tractable manner, by the DNNs architecture. From a field-theory viewpoint this can be stated as
\begin{gather}
\label{Eq:P0fromDelta}
P_0[f] = \int dW P_0(W) \delta[f-z_W]
\end{gather}
at infinite width, where $z_W(x)$ is the output of a DNN with weights $W$ on an input point $x$. The keen reader may be alarmed by the fact that this definition of $P_0[f]$ does not involve the measure $\mu(x)$. However, as shown in \cite{Rasmussen2005}, the norm $\left\Vert f\right\Vert _{K}^{2}$ (called the RKHS norm), and therefore  Eq. \eqref{eq:PDF_nodata_propto} are in fact the same for any two probability measures with identical support. 
\begin{figure}[t]
    \centering
    \includegraphics[width=\linewidth]{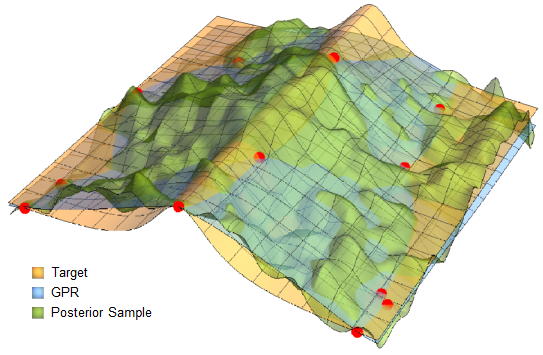}
    \caption{{\bf A physical picture of supervised deep learning.} The output of the DNN, as a function of input data, can be seen as an elastic membrane (surface) which relaxes to its equilibrium distribution during training. In this steady state it fluctuates (green surface) so to maximize its entropy while minimizing its energy. Its energy consists of a data-term pinning it to its target values (yellow surface) on the training points (red-points). In addition an elastic energy term determined by the DNNs architecture, affects its behavior between the training points. For infinitely over-parameterized DNNs, this elastic energy is quadratic and the average surface (blue surface) can be calculated analytically, up to a large matrix inversion, using Gaussian Processes regression.}
    \label{fig:fluctuating_membrane}
\end{figure}

Performing Bayesian inference in the context of GPs means conditioning Eq. \eqref{eq:PDF_nodata_propto} using Bayes' theorem and assuming Gaussian noise with amplitude $\sigma^2$ on our target function ($g(x)$). This yields the additional factor 
\begin{gather}
	\label{eq:PDF_withdata_propto}
    P\left[f\right]\propto e^{-\frac{1}{2}\left\Vert f\right\Vert _{K}^{2}-\frac{1}{2\sigma^2}\sum_{i=1}^{N}\left(f(x_i)-g(x_i)\right)^2}
\end{gather}

It can be checked that the expectation value of $f(x_*)$ under the above probability yields Eq. \eqref{Eq:GPPred}. 

Notably, by taking into account Eq. \eqref{Eq:P0fromDelta}, the above expression coincides with that obtained via the NNSP correspondence Eq. \eqref{Eq:Langevin_PDF} in the infinite over-parameterization limit where $P_{nd}[f]=P_0[f]$ for MSE loss and a suitably chosen $K_{xx'}$. In the NNSP context, the data-term came out quadratic when training using MSE loss and more generally it could be replaced with a general loss function $\mathcal{L}\left[f\right]$, so the DNNs predictive distribution becomes

\begin{gather}
	\label{eq:PDF_generalloss_propto}
    P\left[f\right]\propto e^{-\frac{1}{2}\left\Vert f\right\Vert _{K}^{2}-\frac{1}{2\sigma^2}\mathcal{L}\left[f\right]}
\end{gather}
where $K$ in this context is the kernel of the DNN trained with no data. Though not necessarily Gaussian, this expression can still be treated using mean-field or perturbative approaches. A more detailed treatment of different loss functions, most notably as cross-entropy loss, is left for future work.

Denoting $S[f]=\frac{1}{2}\left\Vert f\right\Vert _{K}^{2}+\frac{1}{2\sigma^2}\mathcal{L}\left[f\right]$ (the "Information Hamiltonian", in IFT terminology), Eq. \eqref{eq:PDF_withdata_propto} gives rise to the partition function 

\begin{align}
	Z[\alpha]=\intop \D f e^{-S[f] + \int dx \alpha(x) f(x)}
\end{align}

where $\int dx \alpha(x) f(x)$ is a source term used to calculate cumulants of $P[f]$, and specifically the average prediction of the network:

\begin{align}
\label{eq:logZvarDer}
g^*(x_*) = \left.\frac{\delta \log(Z[\alpha])}{\delta\alpha(x_*)}\right|_{\alpha=0} = \frac{1}{Z[0]} \int \D f \cdot f(x_*) e^{-S[f]}
\end{align}
where $\delta/\delta\alpha$ stands for functional derivative. As shown visually in Fig. \ref{fig:fluctuating_membrane}, this expression leads to a tangible physical picture of how DNNs learn. Their output as a function of the input can be seen as a fluctuating elastic membrane over input space which, in the highly over-parameterized limit, is in its linear elastic regime. The training data appears as isolated points at which this membrane is pinned down to certain value by (loss dependent) springs whose constant is proportional to $1/\sigma^2$-- the  inverse of the noise on the gradients during training. The membrane then interpolates and extrapolate between these pinning points in a way which, on average, minimizes its elastic energy. This elastic energy differs considerably from that of physical membrane and in particular has a non-local dependence on the shape of the membrane. Different DNNs correspond to different elastic energies. Finite networks entail non-linear corrections to the elastic energy which may be beneficial for learning in the case of CNNs \cite{Gadi2020}.  

\subsection{Predictions in the grand-canonical ensemble}

As mentioned, in order to calculate the learning curve one needs to calculate quantities like $\langle g^*\rangle_{D_N}$ and $\langle {g^*}^2\rangle_{D_N}$. These averages involve multidimensional integrations over all possible datasets. To facilitate their computation we adopt the approach of \citep{Malzahn2001} and instead consider a related quantity given by the Poisson averaging of the former

\begin{align}
\label{Eq:poiAv}
\langle ... \rangle_{\eta} &= e^{-\eta} \sum^{\infty}_{n=0} \frac{\eta^n}{n!} \langle ... \rangle_{D_n} 
\end{align}

where $...$ can be any quantity, in particular $g^*$ and $g^{*2}$. This average can be thought of as a grand-canonical ensemble, though a non-standard one since we average the observables and not the partition function. Taking $\eta = N$ means we are essentially averaging over values of $N$ in an $\sqrt{N}$ vicinity of $N$. This means that as far as the leading asymptotic behavior is concerned, one can safely exchange $N$ and $\eta$ as the differences would be sub-leading. We therefore focus on calculating the grand-canonical DAEE, $\langle \left\Vert g^* - g \right\Vert^2 \rangle_{\eta}$. In App. \ref{Appendix:PoisAvgDemo} we compare learning curves as a function of $N$ and $\eta$ and show that they match very well. 

By using the grand canonical ensemble, averaging over draws from the dataset can be carried out as follows.

First, using the replica trick:

\begin{gather}
\label{Eq:ReplicaTrick}
    \langle g^*(x_*) \rangle_\eta = \lim_{M\rightarrow0}\left. \frac{1}{M} \frac{\delta\langle Z^M\rangle_\eta}{\delta \alpha(x_*)}\right\vert_{\alpha=0}
\end{gather}

where for integer $M$ and assuming that the loss function acts point-wise on the training set $\mathcal{L}[f]=\sum_{i=1}^n\mathcal{L}_f(x_i)$, we have

\begin{align}
\label{Eq:ReplicaTrickZ}
    \langle Z^M \rangle_\eta =& e^{-\eta}\int \prod_{m=1}^M \D f_m \\ \nonumber
    &e^{-\sum_{m=1}^M \left( \frac{1}{2}\left\Vert f_m\right\Vert _{K}^{2}
    - \int dx \alpha f_m\right)
    + \eta \int d\mu_x e^{-\frac{\sum_{m=1}^M \mathcal{L}_{f_m}(x)}{2\sigma^2}}} \nonumber
\end{align}
As shown in App. \ref{Appendix:GprFieldTheoryPred}, a Taylor expansion in $\eta$ of the above r.h.s. yields the $\langle... \rangle_{\eta}$ averaging appearing on the l.h.s. 

Second, we notice that the main benefit of Eq. \eqref{Eq:ReplicaTrick} and Eq. \eqref{Eq:ReplicaTrickZ} over Eq. \eqref{Eq:GPPred} is that it allows for a controlled expansion in $1/\eta$. At large $\eta$ (or similarly large $N$) we expect the fluctuations in $f_m(x)$ to be small and centered around $g(x)$. Indeed, such a behavior is encouraged by the term multiplied by $\eta$ in the exponent. We can therefore systematically Taylor expand the inner exponent

\begin{align}
\label{eq:ExpTaylor}
e^{-\frac{\sum_{m=1}^M \mathcal{L}_{f_m}(x)}{2\sigma^2}} &= 1-\frac{\sum_{m=1}^M \mathcal{L}_{f_m}(x)}{2\sigma^2} \\ \nonumber
&+\frac{1}{2}  \left[\frac{\sum_{m=1}^M \mathcal{L}_{f_m}(x)}{2\sigma^2}\right]^2 + ...      
\end{align}

and each term will yield a higher order of $\langle g^*(x_*) \rangle_\eta$ in $1/\eta$. 

\subsection{EK as a free theory}

Notably, so far the choice of a loss function was largely arbitrary. The advantage of choosing MSE loss, $\mathcal{L}_f(x) = (f(x)-g(x))^2$, is that $P[f]$ also becomes a GP, or equivalently has a quadratic action. From now on we shall focus on MSE loss.

Aiming for standard pertubative calculations, we wish to perform diagrammatic calculations w.r.t a free quadratic theory. 
Expanding Eq. \eqref{eq:ExpTaylor} to first order and substituting in Eq. \eqref{Eq:ReplicaTrickZ} we obtain

\begin{gather}
\label{eq:Z_EK}
\langle Z^M\rangle_\eta = Z_{EK}^M + O\left(1/\eta^2\right) \\ \nonumber
Z_{EK}\left[\alpha\right]= \int\D f e^{-S_{EK}[f] + \int dx \alpha(x) f(x)}
\end{gather}
where $S_{EK}[f]=\frac{1}{2}\left\Vert f\right\Vert _{K}^{2}+\frac{\eta}{2\sigma^{2}}\int d\mu_{x}(f(x)-g(x))^2$, which is quadratic in $f$ and therefore induces a Gaussian field.

Substituting Eq. \eqref{eq:Z_EK} in Eq. \eqref{Eq:ReplicaTrick} we get

\begin{align}
\left\langle g^*(x_*)\right\rangle_\eta &= \left. \frac{\delta\log( Z_{EK})}{\delta \alpha(x_*)}\right\vert_{\alpha=0}+ O\left(1/\eta^2\right)
\\ \nonumber 
&=\arg\min\left[S_{EK}[f]\right] + O\left(1/\eta^2\right) \\ \nonumber
&=g^*_{EK,\eta}(x_*)+ O\left(1/\eta^2\right)
\end{align}
where the second equality is due to the fact that for Gaussian distributions the expectation value coincides with the most probable value, and the third equality is due to \cite{Rasmussen2005}, with the subtle change that $N$ is being replaced by $\eta$.

Let us denote by $\langle\ldots\rangle_0$ the free-theory average, that is an average w.r.t $Z_{EK}$. We therefore get $\langle f\rangle_0=g^*_{EK,\eta}\neq0$, meaning that our free theory, though Gaussian, is not centered.  

The correlations of the free theory are

\begin{gather}
 \mathrm{Cov}_0\left[f(x),f(y)\right] = \sum_{i}\left(\frac{1}{\lambda_{i}}+\frac{\eta}{\sigma^{2}}\right)^{-1}\phi_{i}\left(x\right)\phi_{i}\left(y\right)
\end{gather}

where again, $\lambda_i$ and $\phi_i$ are the eigenvalues and eigenfunctions of the kernel.

\subsection{Next order corrections}
We now wish to perform pertubative calculations w.r.t to the free (Gaussian) EK theory, and obtain a sub-leading (SL) correction for the EK result in the inverse dataset size:

\begin{align}
\label{eq:g_nextOrderCorrection}
\langle g^{*}(x_*) \rangle_{\eta} &= g^*_{EK,\eta}(x_*) + g^*_{SL,\eta}(x_*) + O(1/\eta^3) 
\end{align}

Expanding Eq. \eqref{eq:ExpTaylor} to second order, substituting in Eq. \eqref{Eq:ReplicaTrickZ} and keeping only $ O\left(1/\eta^2\right)$ terms, the calculation can be carried using Feynman diagrams w.r.t to the free EK Gaussian theory. Leaving the details to appendix \ref{App:NextOrderCorrections}, the sub-leading correction is

\begin{align}
\label{eq:g_SL_implicit}
    g^*_{SL,\eta}(x_*)=& \frac{\eta}{\sigma^{4}}\intop d\mu_{x}\left( g^*_{EK,\eta}\left(x\right)-g\left(x\right)\right) \\ \nonumber
    &\mathrm{Cov}_{0}\left[f\left(x\right),f\left(x\right)\right]\mathrm{Cov}_{0}\left[f\left(x\right),f\left(x_*\right)\right] 
\end{align}
or explicitly
\begin{align}
\label{eq:g_SL_explicit}
    g^*_{SL,\eta}(x_*) &= \\ \nonumber -\frac{\eta}{\sigma^{4}}&\sum_{i,j,k} \Lambda_{i,j,k}
    g_{i}\phi_{j}\left(x_{*}\right) \intop d\mu_{x}\phi_{i}\left(x\right)\phi_{j}\left(x\right)\phi_{k}^{2}\left(x\right) \\ \nonumber
    \Lambda_{i,j,k} &= \frac{\frac{\sigma^{2}}{\eta}}{\lambda_{i}+\frac{\sigma^{2}}{\eta}}\left(\frac{1}{\lambda_{j}}+\frac{\eta}{\sigma^{2}}\right)^{-1}\left(\frac{1}{\lambda_{k}}+\frac{\eta}{\sigma^{2}}\right)^{-1}
\end{align}

As shown App. \ref{Appendix:GprFieldTheoryPred}, similar expressions for $\langle g^{*2} \rangle_{\eta}$ are obtained using two replica indices. Interestingly we find that $\langle g^{*2} \rangle_{\eta} = \langle g^* \rangle_{\eta}^2 + O(1/\eta^3)$. Hence, up to $O(1/\eta^3)$ corrections, the averaged MSE error is $(\langle g^*(x_*)\rangle_{\eta}-g(x_*))^2$ integrated over $x_*$. Since the variance of $g^*$ came out to be $O(1/\eta^3)$ one finds that $g^*-g$, which is $O(1/\eta)$, is asymptotically much larger than its standard deviation. This implies self averaging at large $\eta$, or equivalently that our dataset-averaged results capture the behavior of a single fixed dataset. 

Equations \eqref{eq:g_nextOrderCorrection}, \eqref{eq:g_SL_explicit} and their application to the calculation of the grand-canonical DAEE are one of our key results. They provide us with closed expressions for the DAEE as a function of $\eta$, namely the fixed-teacher learning curve. They hold without any limitations on the dataset or the kernel and yield a variant of the EK result along with its sub-leading correction. From an analytic perspective, once $\lambda_i$ and $\phi_i(x)$ are known, the above expressions provide clear insights to how well the GP learns each feature and what cross-talk is generated between features due to the second sub-leading term. Notably for the renormalized NTK introduced below, the number of non-zero $\lambda_i$'s is finite, and so the above infinite summations reduce to finite ones. This makes these expressions computationally superior to directly performing the matrix-inversion in Eq. \eqref{Eq:GPPred} along with an $N-$dimensional integral involved in dataset-averaging. In addition, having the sub-leading correction allows us to estimate the range of validity of our approximation by comparing the sub-leading and leading contributions, as we shall do for the uniform case below.

\section{Uniform datasets}
\label{Sec:Uniform}
To make Eq. \eqref{eq:g_SL_explicit} interpretable, $\phi_i(x)$ and $\lambda_i$ are required. This can be done most readily for the case of datasets normalized to the hypersphere ($\left\Vert x_n\right\Vert=1$) with a uniform probability measure and rotation-symmetric kernel functions. By the latter we mean $K_{x,x'}=K_{Ox,Ox'}$ for any orthogonal matrix $O$ with the same dimension as the inputs. Although beyond the scope of the current work, obvious extensions to consider are datasets which are uniform only in a sub-space of $x$ and/or small perturbations to uniformity.

Importantly, both NNGP and NTK associated with any DNN with a fully connected first layer and weights initialized from a normal distribution, has the above symmetry under rotations (see App. \ref{Appendix:NNGP_RotInv}). It follows that such a kernel can be expanded as $K_{x,x'} = \sum_n b_n (x \cdot x')^n$. An additional corollary \cite{dotProdKer} is that its features are hyperspherical harmonics ($Y_{lm}(x)$) as these are the features of all dot product kernels. Hyperspherical harmonics are a complete and orthonormal basis w.r.t a uniform probability measure on the hypersphere. Note that this implies a non-standard normalization for the $Y_{lm}$s in this context, as they are usually normalized w.r.t Lebesgue measure. For each $l$ these can be written as a sum of polynomials in the input coordinates of degree $l$. The extra index $m$ enumerates an orthogonal set of such polynomials (of size $\operatorname{deg}(l)$). For a kernel of the above form the eigenvalues are independent of $m$ and given by \cite{dotProdKer}
\begin{align}
\label{Eq:generalEigen}
\lambda_l &= \frac{\Gamma\left(\frac{d}{2}\right)}{\sqrt{\pi}\cdot2^{l}}\sum_{s=0}^{\infty}b_{2s+l}\frac{\left(2s+l\right)!}{\left(2s\right)!}\frac{\Gamma\left(s+\frac{1}{2}\right)}{\Gamma\left(s+l+\frac{d}{2}\right)}
\end{align}

For ReLU and erf activations, the $b_n$'s, can be obtained analytically up to any desirable order \cite{Saul2009}. Thus one can semi-analytically obtain the eigenvalues up to any desired accuracy. For the particular case of depth 2 ReLU networks with no biases, we report in App. \ref{Appendix:Insights} closed expression where the above summation can be carried out analytically for the NNGP and NTK kernels. However, as we shall argue soon, it is in fact desirable to trim the NTK in the sense of cutting-off its Taylor expansion at some order $m$, resulting in what we call the renormalized NTK. For such kernels, which would be our main focus next, Eq. \eqref{Eq:generalEigen} can be seen as a closed analytical expression for the eigenvalues. 

Interestingly, for any dot-product kernel and uniform data of dimension $d$ on the hypersphere, there is a universal bound given by $\lambda_l \leq K_{x,x}/\deg(l) \approx O(d^{-l})$, where $K_{x,x}$ is a constant in $x$. 
Indeed, $K_{x,x}= \sum_{lm} \lambda_{l} = \sum_l \deg(l) \lambda_l$. The degeneracy ($\deg(l)$) is fixed from properties of hyper spherical harmonics, and equals $\deg(l) =\frac{2l+d-2}{l+d-2} \binom{l+d-2}{l}$ \cite{SphHarRef} which goes as $O(d^l)$ for $l \ll d$. This combined with the positivity of the $\lambda_l$'s implies the above bound. 

Expressing our target in this feature basis $g(x) = \sum_{l,m} g_{lm} Y_{lm}(x)$, Eq. \eqref{eq:g_SL_explicit} simplifies to 

\begin{align}
\label{Eq:gNoise}
g^{*}_{SL,\eta}(x_{*}) &= - \sum_{l,m}\frac{\eta^{-1} \lambda_l C_{K,\sigma^2/\eta}}{(\lambda_l+\sigma^2/\eta)^2}g_{lm} Y_{lm}(x_*) 
\end{align}
where $C_{K,\sigma^2/\eta} = \sum_{l} \deg(l) (\lambda_{l}^{-1} + \eta/\sigma^2)^{-1}$ and notably cross-talk between features has been eliminated at this order since $\sum_m Y_{lm}^2(x)=\deg(l)$ is independent of $x$, yielding $\sum_{\tilde{m}}\int d\mu_x Y_{lm}(x) Y_{l'm'}(x)Y^2_{
\tilde{l}\tilde{m}}(x) = \deg(\tilde l) \delta_{ll'}\delta_{mm'}$. 

By splitting the sum in $C_{K,\sigma^2/\eta}$, to cases in which $\lambda_l < \sigma^2/\eta$ and their complement, one has the bound $C_{K,\sigma^2/\eta} < \#F \sigma^2/\eta + \sum_{lm | \lambda_l < \sigma^2/\eta} \lambda_l$, where $\#F$ is the number of eigenvalues such that $\lambda_l>\sigma^2/\eta$. Thus for kernels with a finite number of non-zero $\lambda_i$'s (as the renormalized NTK introduced below), and for large enough $\eta$, $\#F$ becomes the number of non-zero eigenvalues and $C_{K,\sigma^2/\eta}=\#F\sigma^2 /\eta$ has a $\eta^{-1}$ asymptotic. This illustrates the fact that the above terms are arranged by their orders in $\eta$. 

We can use Eq. \eqref{Eq:gNoise} to understand the validity of the EK result. We therefore look for sufficient conditions for $g^*_{EK,\eta}\gg g^*_{SL,\eta}$ to hold.
By a term-wise comparison, for some $l$ we obtain $C_{K,\sigma^2/\eta}\ll \eta(\lambda_l+\sigma^2/\eta)$ which holds for $C_{K,\sigma^2/\eta} \ll \sigma^2$.
For trimmed kernels, this yield $\#F \ll \eta$. Notably it means that the original non-trimmed NTK cannot be analyzed perturbatively, since with $\sigma^2=0$, $\#F$ becomes infinite. In the next section we tackle this issue. 

\section{Generalization in the noiseless case and the renormalized NTK}
\label{Sec:RG}

The correspondence between DNNs trained in the NTK regime and GPR using NTK implies noiseless GPR ($\sigma^2=0$) for which the perturbative analysis carried in previous sections fails. Here we show that the fluctuations of $f$ associated with small $\lambda_l$s can be traded for noise on the fluctuations of $f$ associated with large $\lambda_l$s, thereby making our perturbative analysis applicable. As shown in the previous section, for uniform datasets, the smaller $\lambda_l$s correspond to higher spherical harmonics (higher $l$) and hence have higher oscillatory components. We argue that these higher oscillatory modes can be marginalized over in a controlled manner to generate both noise and corrections to the large $\lambda_l$s. This is very much in spirit of the renormalization group technique, wherein high oscillatory modes are integrated over to generate changes (renormalization) of some parameters in the probability distribution of the low oscillatory modes.  

We begin by defining a set of renormalized NTKs. As argued, an NTK of any fully-connected DNN can be expanded as $K_{x,x'} = \sum_{q=0}^{\infty} b_q (x\cdot x')^q$. The renormalized NTK at scale $r$ is then simply $K^{(r)}_{x,x'} = \sum_{q=0}^{r} b_q (x\cdot x')^q$. Harmoniously with this notation we denote the prediction of GPR with the original kernel as $g_{\infty}^*$. Our claim is that GPR with $K$ and a noise of $\sigma^2$ can be well approximated by GPR with $K^{(r)}$ and noise $\sigma^2+\sigma_r^2$ (where $\sigma_r^2=\sum_{q=r+1}^{\infty} b_q$), for sufficiently large $r$. Specifically, our claim is that the discrepancy between the original vs. truncated GPR predictions scales as $O(\sqrt{N} d^{-(r+1)/2}/K_{x,x})$, where $d$ is the effective data-input dimension. Importantly, as can be inferred from Eq. \eqref{Eq:generalEigen}, the renormalized NTK at scale $r$ has zero eigenvalues for all spherical Harmonics with $l>r$, as well as modified eigenvalues for spherical harmonics with $l\leq r$ (compared to the non-truncated NTK). Thus, as advertised, these high Fourier modes have been removed from the problem in exchange for a renormalized theory with a modified low energy spectrum, and augmented noise. In a related manner, trimming the Taylor expansion after $(x \cdot x')^r$ effectively reduces our angular resolution and coarse grains the fine angular features captured by these spherical Harmonics with $l>r$. 

To justify this approximation we consider the difference matrix $A_{nm}=K_{x_n,x_m}-K^{(r)}_{x_n,x_m}$, given a dataset $\{x_n\}_{n=1}^N$ drawn from a uniform distribution on a hypersphere of dimension $d$. The terms $b_q (x_n \cdot x_m)^{q}$ scale roughly as $d^{-q/2}$ (see App. \ref{AppRG} for a more accurate expression) due to the tendency of random vectors in high dimensions to be orthogonal. Consequently the above difference diminishes very quickly with $r$. Notably this also applies for the entries of $K_{x_*,x_n}-K^{(r)}_{x_*,x_m}$, provided $x_*$ is a test point and not a train point. In contrast, the diagonal part of $A$ is $A_{nn}=\sigma_{r}^2$ and may diminish more slowly depending on the coefficients $b_{q>r}$. Upon neglecting $K_{x_*,x_n}-K^{(r)}_{x_*,x_m}$ and the off-diagonal elements of $A$, one finds that Eq. \eqref{Eq:GPPred} with these two GPRs yields identical predictions. As shown in App. \ref{AppRG}, these neglected off-diagonal elements yield a discrepancy which scales as $\sqrt{N} d^{-(r+1)/2}$. Consequently, the MSE error between the two GPRs should scale as $N$ times an exponentially small factor ($d^{-r-1}$). This scaling with $N$ should saturate when the accuracy is nearly perfect since then the predictions remain largely constant as $N$ is increased. 

Focusing back on the question of how to tackle noiseless GPR, we thus find that as long as the $b_q$'s decays slowly enough with $q$, then at any finite $N$ we can choose a large enough $r$ such that two desirable properties are maintained: A. The discrepancy between the GPRs is small and B. $\sigma_r^2$ is large enough to ensure convergence to our perturbative analysis. The required slow decay of $b_q$ is harmonious with the intuition that DNNs should be initialized at the edge of chaos \cite{Schoenholz2016} where the output of the network has a fine and multi-scale sensitivity to small changes in the input. As $K_{x,x'}$ is the correlation of two outputs with inputs $x$ and $x'$, having a power law decaying $b_q$ implies such fine and multi-scale sensitivity. Establishing relations between good initialization and effectiveness of our renormalized NTK is left for future work.    

We have tested the accuracy of approximating noiseless NTK GPR with renormalized NTK GPR with the appropriate $\sigma_r^2$, both on artificial datasets (see next section) and on real world dataset such as CIFAR10 (see app. \ref{Appendix:CompRNTK}). In both cases we found an excellent agreement between the two GPRs for $r$'s as small as $3$ and $4$.


\section{Generalization in the NTK regime} 
\label{Sec:ConcreteExample}

\begin{figure*}[t]
     \centering
     \includegraphics[width=\linewidth]{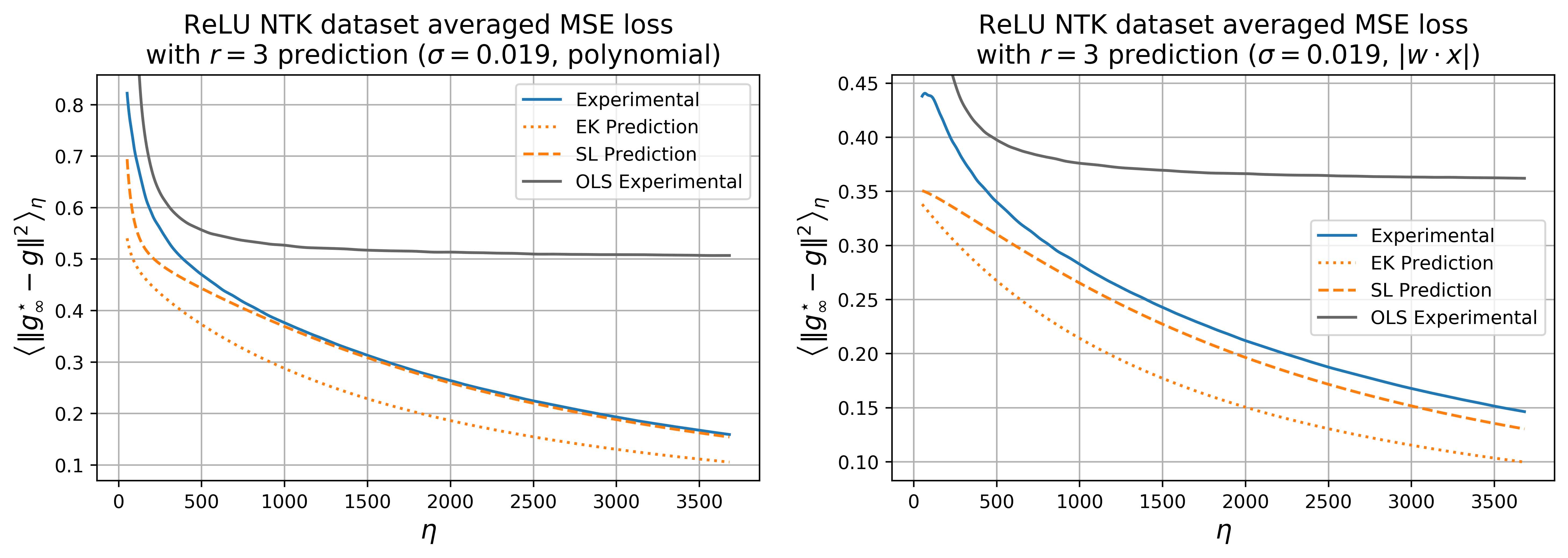}
     \caption{The experimental learning curves (solid lines) for a depth 4 ReLU networks trained in the NTK regime on different target functions on a $d=50$ hypersphere are shown along with our analytical predictions for the leading (dotted line) and leading plus sub-leading behavior (dashed line). Left panel shows the results for a second order polynomial in the input, and the right panel show results for the function $|w\cdot x|$ (where $w$ is a random vector of norm $\sqrt{d}$) which cannot be expressed as a finite linear combinations of eigenfunctions. The learning curves of ordinary least squares (OLS) on the same regression tasks are provided to help compare the performance of GPs with simpler regression methods.}
     \label{Fig:learning_curve_prediction}
\end{figure*}

Collecting the results of all the preceding sections, we can obtain a detailed and clear picture of generalization in fully connected DNNs trained in the NTK-regime on datasets with a uniform distribution normalized to some hypersphere in input space. 
We begin with a qualitative discussion and consider some renormalized NTK at scale $r$. From Sec. \ref{Sec:Uniform}, we have that the features of this kernel are hyperspherical harmonics and that $\lambda_l$ scales as $d^{-l}$. We also recall that $Y_{lm}$ is a polynomial of degree $l$ and that all the hyperspherical harmonics up to degree $l$ span all polynomials on the hypersphere with degree up to $l$. Examining Eq. \eqref{Eq:gNoise} we find that features with $\lambda_{l} \gg \sigma^2/\eta$ are learnable and via the above scaling we find that we learn polynomials of degree $O(\log(\eta/\sigma^2)/\log(d))$ or less. In particular, a function like parity, which is a polynomial of degree $d$ is very hard to learn whereas a linear function is the easiest to learn. Thus, despite having infinitely more parameters than data-points (due to infinite width) and despite being able to span almost any function (due to the richness of the kernel's features), the DNN here avoids overfitting by having a strong bias towards low degree polynomials.  

To make more quantitative statements we now focus on a specific setting. We consider input data in dimension $d=50$ and a scalar target function $g(x)=\sum_{l=1,2;m} g_{lm} Y_{lm}(x)$ such that the vectors $(g_{l,1},g_{l,2},\ldots,g_{l,\deg(l)})^T$ for $l=1,2$ are drawn from a uniform measure on the $\deg(l)$-sphere of radius $1/\sqrt2$. We generate several toy datasets $D_N$ consisting of $N$ data points ($x_n$) uniformly distributed on the hypersphere $S^{d-1}$ and their corresponding targets ($g(x_n)$). We consider the GP equivalent to training a fully-connected DNN consisting of 4 layer with ReLU activations and width $W$ which we initialize with variance ($\sigma^{2}_{w}=\sigma^{2}_{b}=1/d$) for the input layer and ($\sigma^{2}_{w}=\sigma^{2}_{b}=1/W$) for the hidden layers (see for instance \citep{Lee2019} App. C and App. E for how to compute the kernel. Notice there is a factor of $1/W$ between our convention for $\sigma_{w}^2$ and \citep{Lee2019}). To be in the NTK correspondence regime we consider training such a network at vanishing learning-rate, MSE loss, and with $W\gg N$. One then has that the predictions of the DNN  are given by GPR with $\sigma^2=0$ and the $K$ given by the NTK kernel \cite{Jacot2018} (To be more precise, \cite{Jacot2018} predict correspondence with GPR up to a random initialization factor, so to get exact match with GPR one would also need to average over initialization seeds. Recent research \cite{Lee2019} suggests this caveat can be avoided under some conditions).

For each such DNN we obtained the expected MSE loss $\left\lVert g_{\infty}^*-g\right\rVert^2$ of GPR with the NTK kernel by numerical integration over $x_*$.   Repeating this process multiple times we obtained the DAEE for $N=1,2,\dots,N_{\max}$ with a relative standard error of less then 5\% (this typically required averaging over 10 datasets). For direct comparison with our prediction of the learning curve, we computed the Poisson averaged learning curve $\langle\left\lVert g_{\infty}^*-g\right\rVert^2\rangle_{\eta}$ in accordance with Eq. \eqref{Eq:poiAv}, neglecting the terms $n>N_{\max}$. We restricted ourselves to $\eta_{\max}\le N_{\max}-5\sqrt{N_{\max}}$ to make tail effects negligible. Notably the Poisson averaging makes the final statistical error negligible relative to the discrepancies coming from our large $\eta$ approximations (see \ref{Appendix:PoisAvgDemo}). To make it easier to appreciate the power of GPs over simpler regression models we also provide the Poisson averaged DAEE for ordinary least square method (OLS) as a yardstick.

To pick the renormalization scale $r$ we must consider two factors, on the one hand we want the discrepancy between the renormalized and regular NTK to be small, this scales as $O(\sqrt{N} d^{-(r+1)/2}/K_{x,x})$. On the other hand we want the effective noise $\sigma^2_{r}$ to be as large as possible to assure the accuracy of the prediction. We found that $r=3$ strikes a good balance for the range of $N$ values used in the experiment, but $r=4,5,6$ also produced adequate predictions since $\sigma^2_{r}$ shrinks slowly with $r$ for the architecture used.

Our analytical expressions following Eq. \eqref{Eq:generalEigen} combined with known results \cite{Jacot2018,Saul2009} about the Taylor coefficients ($b_n$) yield $\lambda_0,...,\lambda_3=\{3.19,7.27\cdot10^{-3},5.98\cdot10^{-6},1.62\cdot10^{-7}\}$ and $\sigma_r^2 = 0.018$. Since $\lambda_0,\lambda_1 \gg \sigma^2/\eta \gg \lambda_{2},\lambda_{3}$ for $50< \eta < 3500$, $C_{K_r,\sigma^2/\eta}\sigma^{-2} < [\deg(0)+\deg(1)] \sigma^2/\eta + O(\deg(2) 10^{-6})$, thus $C_{K_r,\sigma^2/\eta} \sigma^{-2} \approx 51/\eta$. Thus we expect perturbation theory to be valid for $\eta \gg 50$. At $\eta=1000$ the $l=1$ features are learned well since $\sigma^2/\eta=1.8\cdot10^{-4} \gg \lambda_1$ and the $l=2$ features neglected, at $\eta=1000$ they are learned but suppressed by a factor of about $3$. Had the target contained $l=3$ features, they would have been entirely neglected at these $\eta$ scale. Experimental learning curves along with our leading and sub-leading estimates are shown in Fig. \ref{Fig:learning_curve_prediction} left panel showing an excellent agreement between theory and experiment.

While no actual DNNs were trained in the above experiments, the NTK correspondence means that this would be the exact behavior of a DNN trained in the NTK regime \cite{Jacot2018, Lee2019, Arora2019OnExact}. Furthermore, since our aim was to estimate what the DNNs would predict rather than reach SOTA predictions, we focus on reasonable hyper-parameters but did not perform any hyper-parameter optimization. The complementary case of noisy GPR, which one encounters in the NNSP correspondence, is studied in App. \ref{Appendix:NNSPLearningCurves}.

To demonstrate that our results work with more complex functions we repeated the experiment with a different target function which cannot be expressed as a finite order polynomial. We drew a uniformly distributed vector $w$ on the $(d-1)\textrm{-sphere}$ of radius $\sqrt{d}$ and set the target as $g(x)=|w\cdot x|$. Fig. \ref{Fig:learning_curve_prediction} (right) shows good agreement between theory and experiment here as well.

Lastly we argue that the asymptotic behavior of learning-curve we predict is more accurate than the recent PAC based bounds \cite{AllenZhu2018,CaoB2019,Cao2019}. In App. \ref{App:bounds_comparison} we show a log-log plot of the learning-curves contrasted with a $1/\sqrt{\eta}$ which is the most rapidly decaying bound appearing in those works. It can be seen that such an asymptotic cannot be made to fit the experimental learning-curve with any precision close to ours. 

\section{Application of results to hyper-parameter optimization}
\label{Sec:Hyper-parameterOptimization}

As with most machine learning algorithms, when training a neural
network for a particular task one needs to choose a number of hyper-parameters
such as the network's width at each level $W_l$, depth $L$, variance of weights
at initialization $\sigma_{b}^{2},\sigma_{w}^{2}$, activation function,
and optimizer related parameters such as batch size, learning rate
etc. There are many considerations for hyper-parameter selection such
as training time, memory footprint and optimizer convergence, but
here we will focus on the expected loss of the network. While there
are some accepted heuristics, there is no a priori way to predict
the best preforming architecture other than an expensive
process of trial and error. In this section we introduce a scheme
for picking theoretically advantageous parameters, given minimal information
on the target function and dataset distribution. As recent research
suggests \cite{Belkin2018ReconcilingMM} $W$ should be increased as much as possible
to put the network in the interpolation regime, we will assume that
choice was made. We also note that $\sigma_{b}^{2},\sigma_{w}^{2}$
are typically thought to be related more to convergence issues, for
example via the exploding/vanishing gradient problem, than to the
performance of the network. However as this work as well as \cite{Rahaman2018OnTS}
suggest, these parameters have an important effect on the network
performance by changing the NTK spectrum. 

We suggest the following scenario: we have $N=1000$ data points uniformly
distributed on $S^{9}$. We are also given the spectral weight of
the target in each eigenspace, that is $w_{\ell}^2=\Vert \Pi_\ell(g)\Vert^2$ where $\Pi_\ell$ is the projection operator on the $\ell$ subspace. For the case of uniform measure on the hyper-sphere the projection operator is simply $\Pi_\ell=P_\ell(\langle x\cdot y\rangle)\operatorname{deg}(\ell)$ where $P_\ell$ is the Legendre polynomial of degree $\ell$ and $\operatorname{deg}(\ell)$ is the dimension of the eigenspace, so finding $w_{\ell}^2$ is a much simpler task then finding the $\operatorname{deg}(\ell)$ coefficients of the target (which scale as $d^\ell$) and can be accomplished with a few numeric integrals. In this case we focus on a target with $w{_\ell}^2=\frac{1}{3}(\delta_{1,\ell}+\delta_{2,\ell}+\delta_{3,\ell})$. Given this setting, we would like to find a network architecture with minimal expected error. For computational efficiency reasons we decide to focus on ReLU networks with one hidden layer, so we need to choose four hyper-parameters $\boldsymbol{\sigma}=\left(\sigma_{w_{1}},\sigma_{w_{2}},\sigma_{b_{1}},\sigma_{b_{2}}\right)$.

We present two typical ways used to choose $\boldsymbol{\sigma}$, then propose a better way to do so based our theory. The naive and most prevalent way to choose hyper-parameters is to simply take $\boldsymbol{\sigma}_{\textrm{Typical}}=(\sqrt{2},\sqrt{2},0.05,0.05)$ which roughly correspond to He initialization \cite{HeInit}, a common heuristic for avoiding gradient propagation issues. A more diligent approach would be to draw some random values in the vicinity of $\boldsymbol{\sigma}_{\textrm{Typical}}$, train the network, evaluate the test loss and pick the best preforming hyper-parameters $\boldsymbol{\sigma}_{\textrm{Best}}$. 

Next we suggest a different approach which utilizes our analytical results. We construct a symbolic expression for the expected loss using the formalism outlined in the paper. By taking $\eta=N$ and applying the renormalization scheme with appropriate $r$  we get an estimator for the expected loss $\hat{L}\left(\boldsymbol{\sigma}\right)=\int dx\left\langle \left(f\left(x\right)-g\left(x\right)\right)^{2}\right\rangle$
which we can use to predict the performance of different hyper-parameters
without training the network. Moreover, we can use standard numerical
optimization algorithms to minimize the predicted loss and obtain $\boldsymbol{\sigma}_{\textrm{Optimized}}=\operatornamewithlimits{argmin}_{\boldsymbol{\sigma}}\hat{L}\left(\boldsymbol{\sigma}\right)$.

To test the scheme experimentally, we drew 21 random hyper-parameter proposals $\{\boldsymbol{\sigma}_i\}$ uniformly distributed in the hyper-rectangle defined by $\frac{1}{2}\boldsymbol{\sigma}_{\textrm{Typical}}, \frac{3}{2}\boldsymbol{\sigma}_{\textrm{Typical}}$. We constrained the optimization algorithm to this hyper-rectangle as well to avoid convergence issues in the training procedure, as unconstrained optimization leads to very small values of $\sigma_w$ ($\ll\sqrt{2}$) which in turn lead to vanishing gradients and long training times. We defined networks corresponding to $\boldsymbol{\sigma}_{\textrm{Optimized}}, \boldsymbol{\sigma}_{\textrm{Typical}},\{\boldsymbol{\sigma}_i\}$ and trained each network on the same dataset with full-batch gradient descent and learning rate $1.0$ until the train loss was smaller then $0.1\cdot\sigma^{2}_r$. A summary of the experiment is outlined in Table \ref{Tbl:hyperparameterTable}.


\begin{table}
\centering
\begin{ruledtabular}
\begin{tabular}{lrrr}
                    &  Test & Prediction &   GPR \\
              Worst & 0.413 &      0.406 & 0.382 \\
             Median & 0.313 &      0.316 & 0.319 \\
               Best & 0.175 &      0.198 & 0.214 \\
            Typical & 0.307 &      0.307 & 0.317 \\
          Optimized & 0.078 &      0.110 & 0.141 \\
 \end{tabular}
\end{ruledtabular}
\caption{\label{Tbl:hyperparameterTable} Comparison of the performance of networks trained in the NTK regime using different hyper-parameters ($\boldsymbol{\sigma}$). The Test column shows an estimate of the DNN test loss, the Prediction column the loss as predicted by our learning-curve, and the GPR column an estimate of the dataset averaged expected loss of the corresponding GP. Worst, Median and Best refer to one of 21 networks with random hyper-parameters ranked by test loss. Typical and Optimized refer to networks with $\boldsymbol{\sigma}_{\textrm{Typical}}$ (defined in the text) and the optimal hyper-parameters following our optimization scheme. For more experiment results see App. \ref{Appendix:hyperparameterExperimentResults}}
\end{table}

The results clearly demonstrate the effectiveness of our scheme which reduces the test loss by a factor of 4 relative to the typical hyper-parameter choice and a factor of 2 over the best performing random hyper-parameters. In terms of computational complexity, it took approximately 2.5 hours to train each network using Google's neural tangent package \cite{neuraltangents2020} on a 20 core CPU \footnote{While GPUs are generally faster, fully connected DNNs do not gain the full benefit of GPU parallelism and we expect the computation time would only improve by a factor of $O(1)$} with $W=2^{14}$. In comparison, the time it takes to build and optimize $\hat{L}$ is completely negligible at about 30 seconds. The best random hyper-parameters were found on the fifteenth attempt, so had we stopped then we would have wasted 35 computer hours relative to our scheme and gotten inferior test loss. Note also that we focused on $L=2$ in order to speed up training, which scales exponential with depth, but the optimization procedure is not nearly as sensitive to depth and could have been done for any reasonable $L$. Moreover, increasing the depth would have also enlarged the hyper-parameter space, making random search even less effective. For each network we also experimentally obtained the dataset averaged expected loss using GPR with the associated NTK. The fair agreement between the test loss and dataset averaged expected loss (GPR in Table \ref{Tbl:hyperparameterTable}) further solidifies previous results and demonstrates our claim of self-averaging. 

As expected, the above results required some knowledge of the target function, in particular its spectral weight within each angular momentum space. Alternatively one can capitalize on the fact that our learning curves predictions are quadratic in the target, average them over a target function ensemble, and optimize with respect to this average case. Another option is to consider a min-max optimization scheme in which hyper-parameters are optimized for the worse case target within some domain. The scheme can also be extended to non-uniform datasets and different activation functions as long as some way of computing the eigenvalues is provided.




%

\section{Discussion and Outlook}
\label{Sec:Discussion}

In this work we laid out a formalism based on field theory tools for predicting learning-curves in the NTK and NNSP correspondence regimes. Despite DNNs' black-box reputation, well within the validly range of our perturbative analysis, we obtained very low relative mismatch between our best estimate and the experimental curves, with good agreement extending well into regions with low amounts of data compared to that needed to learn the target. A potential use of such learning curves in hyper-parameter optimization was also demonstrated. 

Central to our analysis was a renormalization-group transformation leading to effective observation noise on the target and to a simpler renormalized quadratic-action/kernel. Notably this RG transformation implied that wide Fully-Connected networks, even ones working on real-world datasets such as CIFAR10, could be effectively described by very few parameters being the noise level and the $O(1)$-first Taylor expansion parameters of the kernel. 

Our analysis provides a lab setting in which deep learning can be understood. In its training phase, DNNs avoid local-minima issues and glassy behavior due to their high over parameterization which makes the optimization problem highly under-constraint \cite{Dauphin2014,Draxler2018,Gadi2020}. As a result, many different solutions or weights which fit perfectly the training data are possible. While each such solution will behave differently on a test point, this arbitrariness does not entail an erratic behavior. The reason is the implicit bias DNNs have towards simple functions. In the case of the NNSP correspondence, a simple function is, by definition, a function that can be generated, up to some small noise, by a large phase-space of weights. 

Simplicity is therefore strongly architecture and dataset dependent. For fully connected DNNs trained in the regime of the NTK or the NNSP correspondences, as well as data uniformly sampled from the hypersphere, simplicity amounts to low order polynomials over that hypersphere. These are the hyper-spherical Harmonics with low $l$, which are the leading eigenfunctions w.r.t. such a uniform measure of a generic kernel associated with a fully connected DNN. As long as the DNN has at least one non-linear layer and biases, depth has only a quantitative effect as it modifies the eigenvalues ($\lambda_l$) but does not change their scale. Generally, the eigenvalues and eigenfunctions vary with architecture and data distribution. Convolutional neural networks (CNNs) require further study, however one can argue on a qualitative level that simple functions would be polynomials with certain spatial hierarchy. Moreover, one expects that qualitative details of this hierarchy would depend on depth as it controls the input-fan-in of the hidden activation in the last CNN layer.   

It seems unrealistic that a purely analytical approach such as ours would describe well the predictions of state-of-the-art DNNs such as VGG-19 trained on a real-world datasets such as ImageNet. Similarly unrealistic is to expect an analytical computation based on thermodynamics to capture the efficiency of a modern car engine or one based on Naiver-Stoke's equations to output a better shape of a wing. Still, scientific experience shows that understanding toy-models, especially rich enough ones, has value. Indeed toy-models provide an analytical lab where theories could be refined or refuted, algorithms could be benchmarked and improved, and wider ranging conjectures and intuitions could be formed. Such models are useful whenever domain knowledge possesses some degree of universality or independence from detail. In converse, when all details matter knowledge is nothing more than a log of all experiences. The fact that DNNs work well in variety of different architecture and data-set settings, suggests that some degree of universality worth exploring is present. Further research would thus tell if the tools and methodologies that have enabled us to comprehend our physical world can help us comprehend the artificial world of deep learning. 

Many extensions of the current work, aimed at approaching real-world settings, can be considered. First and most, much of the recent excitement about DNNs comes from either CNNs or Long Short Term Memory networks (LSTMs). Considering CNNs, while much of our formalism applies, the spectrum of CNN Kernels is more challenging to obtain as their Kernels are less symmetric compared to Fully-Connected DNNs. From similar reasons the RG approach presented here requires a more elaborate trimming of the CNN kernel since the latter would not consist of only powers of dot-products. Furthermore, CNNs trained with SGD show rather large gaps in performance compared to their NNGP or NTK. The culprit here might very well be the finite-width or finite-number-of-channels corrections to the NNGP or NTK priors. Leading finite-width corrections, considered in Ref. \cite{Gadi2020}, amount to adding quartic terms to $P_0[f]$. Those could be dealt with straightforwardly using our perturbation theory formalism. Interestingly, at least for CNNs without pooling, these corrections introduce a qualitative change to the prior, making it reflect the weight-sharing property of CNNs which is lost at the level of the NNGP or NTK \cite{Novak2018,Gadi2020}. Other viable directions are handling richer datasets distributions, extending EK results to the more common cross-entropy loss, applying RG reasoning on finite-width DNNs, and using the above insights for developing DNN-architecture design principals. 


{\it Acknowledgements.} Z.R and O.M acknowledge support from ISF grant 2250/19. Both O.M. and O.C. contributed equally to this work. 

\bibliography{LEGO}

\onecolumngrid
\appendix 

\clearpage
\section{Poisson Averaging Demonstration}
\label{Appendix:PoisAvgDemo}

Here we demonstrate that Poisson averaging has no substantial effect on the learning curve. To this end the figure below shows the experimental learning curve from the main text pre- and post-averaging. It is evident that other than the unintended consequence of eliminating the experimental noise, the averaged learning curve is equivalent to the original for all practical intents.

\begin{figure}[h]
    \centering
    \includegraphics[width=0.7\linewidth]{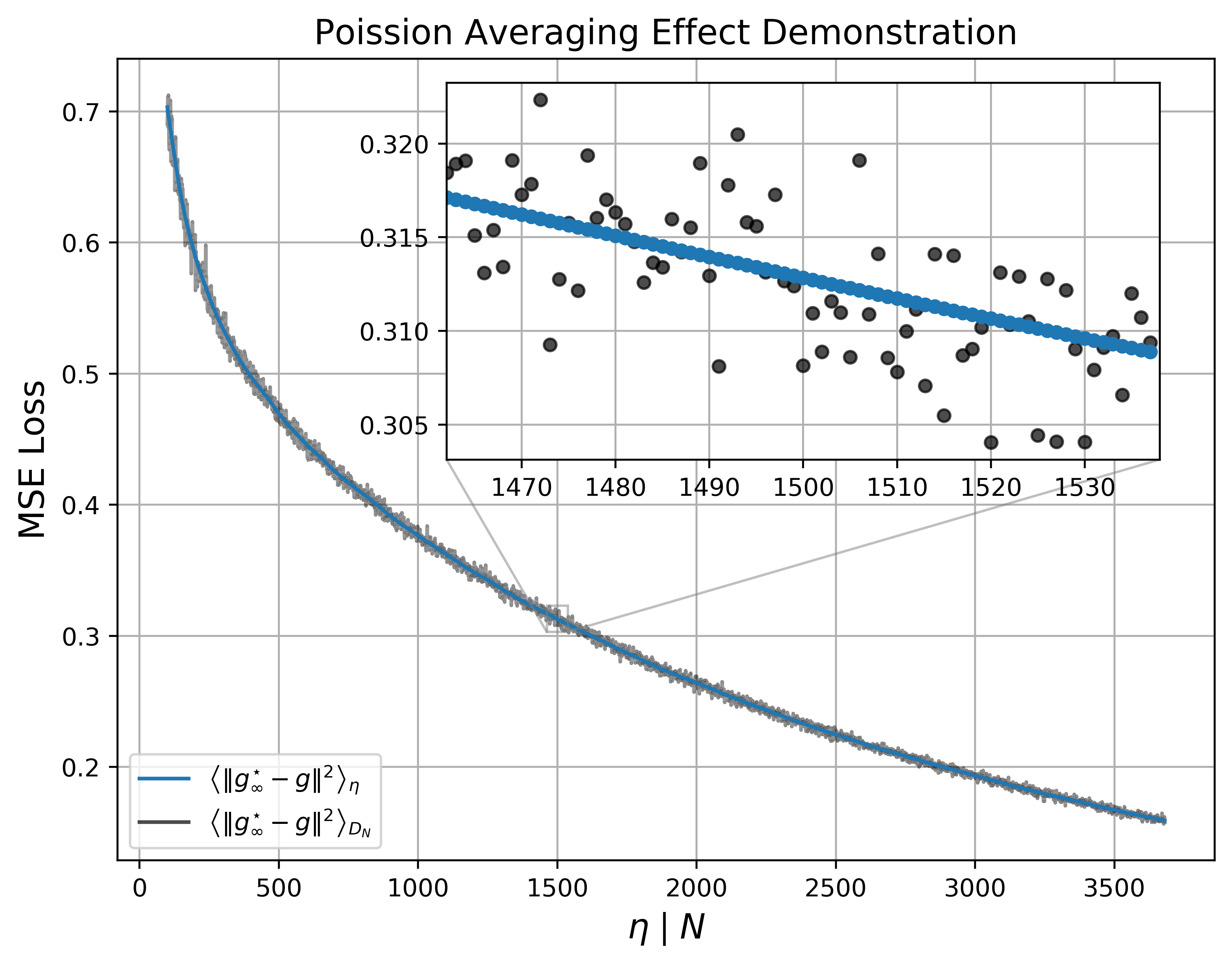}
    \label{Fig:poissAvg}
\end{figure}

\clearpage
\section{Comparison of NTK and Renormalized NTK Predictions on Synthetic and Real World Datasets}
\label{Appendix:CompRNTK}

In section \ref{Sec:ConcreteExample} we used the renormalized NTK as a proxy for the regular NTK, the purpose of this section is to affirm the validity of this approximation. Moreover, while our lack of knowledge of the NTK eigenvalues and eigenfunctions with respect to a non-uniform measure prevents us from predicting learning curves, we would like to show that the renormalized NTK is a valid approximation in this setting as well. 

To this end we used the following procedure. We took the NTK kernel defined in the paper and its associated renormalized kernels at different scales and trained them over the same training set $D_N$. In the figure below (top) the training set and target function were the ones defined in the main text. In the figure below (bottom) $D_N$ consisted of uniform draws without replacements from the cifar-10 training set, standardized and normalized to unit vectors, and the target function was the one-hot encoding of the labels standardized to have zero mean and $K_{x,x}$ variance.  

For each training set we logged the average squared deviation of each renormalized kernel estimation $g^{\star}_r$ from the estimation of the non-renormalized kernel $g^{\star}_\infty$. This is the quantity $\left\Vert g^{\star}_r-g^{\star}_\infty \right\Vert^2$ (where in the cifar-10 case $\Vert\cdot\Vert$ implies both the Euclidean norm in $\mathbb{R}^{10}$ and integration over the input measure, which we approximated by averaging over the cifar-10 test set). We averaged this quantity over different draws of training sets to obtain $\left\langle\left\Vert g^{\star}_r-g^{\star}_\infty \right\Vert^2\right\rangle_{D_N}$. The results show good agreement between $g^{\star}_\infty$ and $g^{\star}_r$ as $r$ is increased.

\begin{figure}[h]
    \centering
    \includegraphics[width=\linewidth]{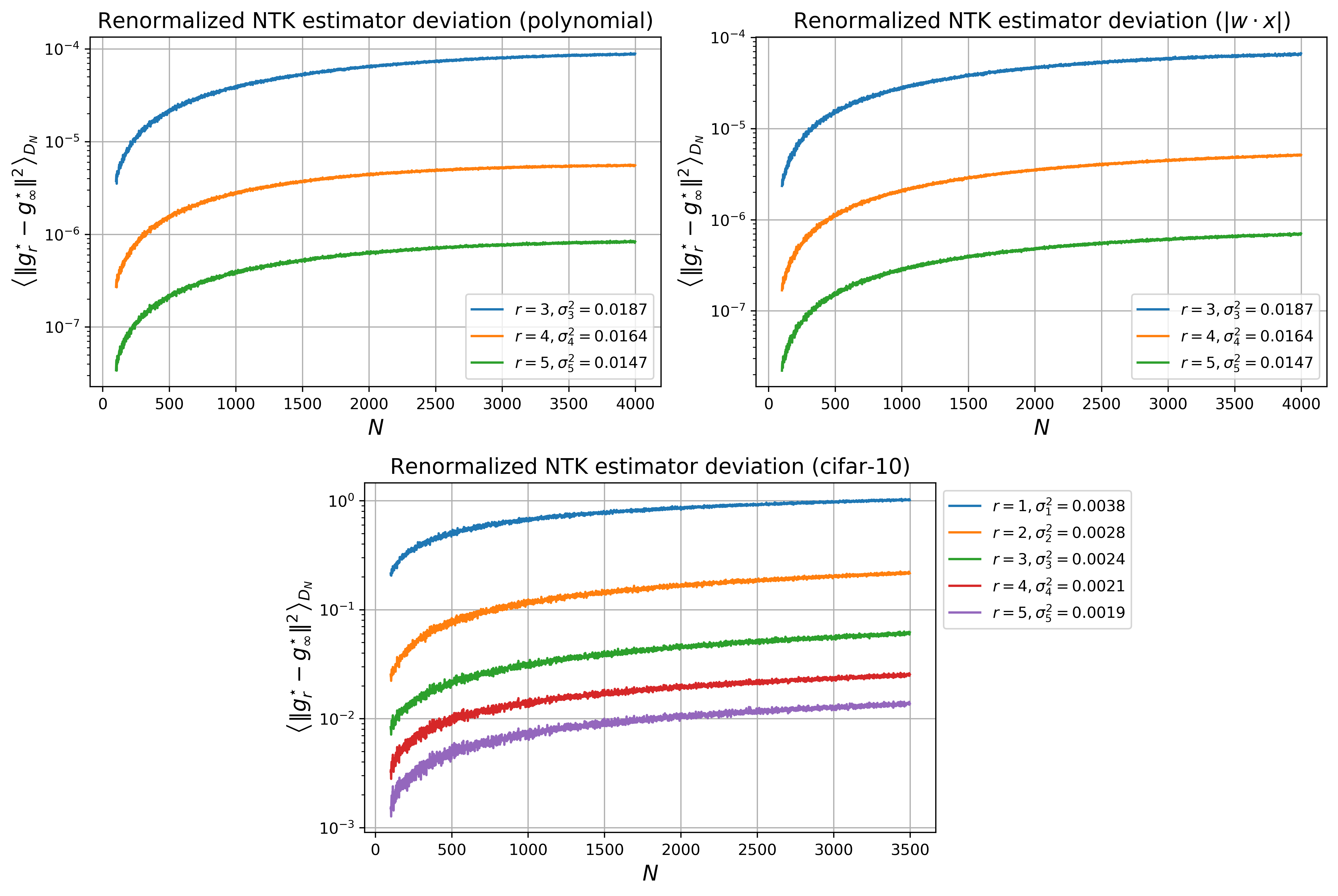}
    \label{Fig:real_data_cmp}
\end{figure}

\clearpage
\section{Learning Curves in the NNSP Protocol} 
\label{Appendix:NNSPLearningCurves}
We report here the results of a similar experiment to the one presented in the main text, but with the NTK kernel replaced with the NNGP kernel as appropriate for the NNSP correspondence. In this case we used a kernel simulating a network with a single hidden layer and $\sigma^{2}_w=1/W, \sigma^{2}_b=0$, and a target function equivalent to the one in the main text. In the NNSP protocol the renormalization group approach is not necessary to introduce noise to the observations, as it comes into play naturally via the temperature dependent fluctuations, so we can choose arbitrary $\sigma^2$. Notwithstanding, the renormalization group approach can aid in analyzing low temperature behavior. 

Notice, in the figure below, that the sub-leading prediction significantly  improves upon the EK prediction. As the inset plot demonstrates, when the dataset size is small the expected error actually increases. Surprisingly, the sub-leading correction manages to capture this behaviour even though the dataset size is small, demonstrating its superiority.

\begin{figure}[h]
    \centering
    \includegraphics[width=0.7\linewidth]{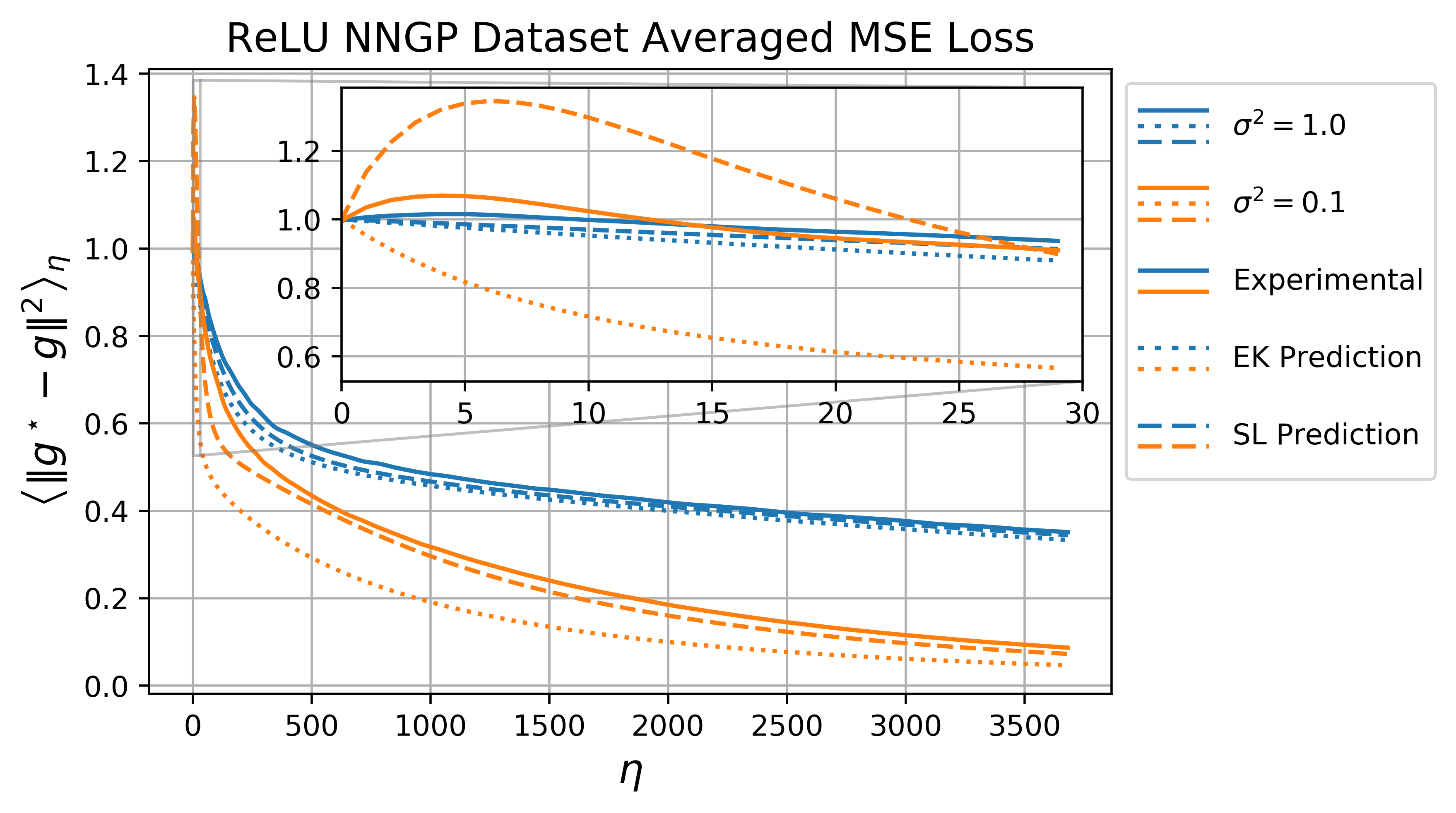}
    \label{Fig:nngp_lc_prediction_plot_prx}
\end{figure}

\clearpage
\section{Comparison with recent bounds} 
\label{App:bounds_comparison}
As mentioned in the main text, various recent bounds, relevant to the NTK regime, have been derived recently. Notwithstanding importance and rigor of these works, their bounds have at best a $1/\sqrt{N}$ asymptotic scaling. The figure below shows that given a functional behavior of the experimental learning curves such a bound cannot be nearly as tight as our predictions.

\begin{figure}[h]
    \centering
    \includegraphics[width=0.7\linewidth]{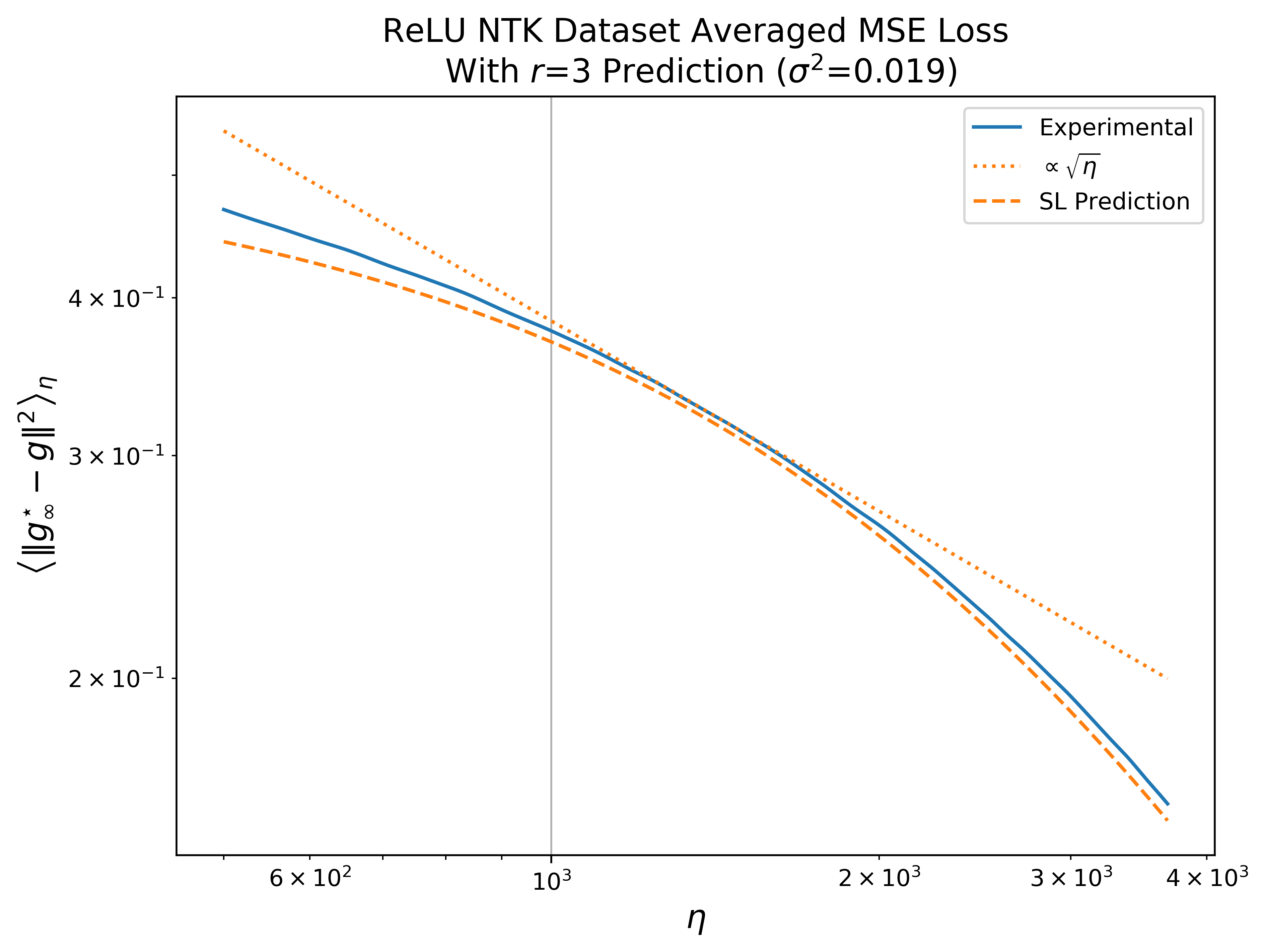}
    \label{Fig:prx_lc_pred_asymp}
\end{figure}

\clearpage
\section{NNGP and NTK are Rotationally Invariant}
\label{Appendix:NNGP_RotInv}

Let us proof that the NNGP and NTK kernels associated with any network whose first layer is fully-connected, are rotationally invariant.
Indeed, let $h_w(x)$ be the output vector of the first layer $[h_w(x)]_i = \phi(\sum_j w_{ij}x_j + b)$ where $x_j$ is the $j$'th component of the input vector $x$. Let $z_{w'}(h)$ be the output of the rest of the network given $h$. The covariance function of NNGPs are defined by \cite{Saul2009}
\begin{align}
K(x,y) &= \int dw dw' P_0(w,w') z_{w'}(h_w(x))z_{w'}(h_w(y))
\end{align}
where $P_0(w,w')$ is a prior over the weights, typically taken to be i.i.d Gaussian for each layer ($P_0(w,w')=P_0(w)P_0(w')$ and $P_0(w) \propto e^{-\sum_{ij} w^2_{ij}/(2\sigma^2)}$). Following this one can show 
\begin{align}
K(Ox,Oy) &= \int dw dw' P_0(w,w') z_{w'}(h_w(Ox))z_{w'}(h_w(Oy)) 
\\ \nonumber
&= \int dw dw' P_0(w,w') z_{w'}(h_{O^Tw}(x))z_{w'}(h_{O^Tw}(y))
\\ \nonumber 
&= \int dw dw' P_0(Ow,w') z_{w'}(h_{w}(x))z_{w'}(h_{w}(y) 
\\ \nonumber
&= \int dw dw' P_0(w,w') z_{w'}(h_{w}(x))z_{w'}(h_{w}(y) = K(x,y)
\end{align}
where the second equality uses the definition of $h_w(x)$, the third results from an orthogonal change of integration variable $w \rightarrow O^Tw$, and the forth is a property of our prior over $w$. Since the NTK relates to the NNGP kernel in a recursive manner (\cite{Jacot2018}), it inherits that symmetry as well.

\clearpage
\section{Notations for the field theory derivation}
\label{Appendix:Notation}

For completeness, here we re-state the notations used in the main-text. 

$x,x',x^{*}$ - Inputs.

$\mu_{x}$ -  Measure on input space.

$K\left(x,x'\right)$ - Kernel function (covariance) of a Gaussian
process. Assumed to be symmetric and positive-semi-definite.

$\phi_{i}\left(x\right)$ - $i$'th eigenfunction of $K\left(x,x'\right)$.
By the spectral theorem, the set $\left\{ \phi_{i}\right\} _{i=1}^{\infty}$
can be assumed to be orthonormal:
\begin{gather*}
\intop d\mu_{x}\phi_{i}\left(x\right)\phi_{j}\left(x\right)=\delta_{ij}
\end{gather*}

$\lambda_{i}$ - $i$'th eigenvalue of $K\left(x,x'\right)$:
\begin{gather*}
\intop d\mu_{x'}K\left(x,x'\right)\phi_{i}\left(x'\right)=\lambda_{i}\phi_{i}\left(x\right)
\end{gather*}

$\Vert\cdot\Vert_{K}$ - RKHS norm:
\begin{gather*}
\Vert\cdot\Vert_{K} = \intop d\mu_{x}d\mu_{x'}f(x)K^{-1}\left(x,x'\right)f(x')
\end{gather*}
If $f\left(x\right)=\sum_{i}f_{i}\phi_{i}\left(x\right)$
then $\Vert f\Vert_{K}=\sum_{i}\frac{f_{i}^{2}}{\lambda_{i}}$
(where $\phi_{i}$ is an orthonormal set). Note that this norm is
independent of $\mu_{x}$ \cite{Rasmussen2005}.

$g\left(x\right)$ - The target function.

$\sigma^{2}$ - Noise variance.

$N$ - Number of inputs in the data-set.

$D_{N}$ - Data-set of size $N$, $D_{N}=\left\{ x_{1},...,x_{N}\right\} $.

$g^*$ - The prediction function.

\clearpage
\section{Phrasing the Problem as a Field Theory Problem}

\subsection{Without Data}
\label{Appendix:ExplicitPathIntegralComputations}
We start by establishing the exact equivalence between a prior of a centered GP and the corresponding partition function. \\
For a kernel function $K$, let us define the partition function
\begin{equation}
\label{Eq:partitionNoData}
    Z[\alpha]=\int\D f\exp\left({-\frac{1}{2}\left\Vert f\right\Vert_K^2}+ \int dx \alpha(x)f(x)\right) 
\end{equation}
Since the RKHS norm is quadratic in $f$, the distribution over the space of functions induced by $Z$ is Gaussian (a GP). Since a GP is determined by is mean and kernel, it is sufficient to show those equalities. 

For the mean we get
\begin{gather}
    \left\langle f(x_*) \right\rangle =\left.\frac{\delta\log(Z[\alpha])}{\delta\alpha\left(x_*\right)}\right|_{\alpha=0}=\\ \nonumber
    \frac{\int \D f\cdot f\left(x_*\right)\exp\left(-\frac{1}{2}\left\Vert f\right\Vert _{K}^{2}\right)}{\int \D f\exp\left(-\frac{1}{2}\left\Vert f\right\Vert _{K}^{2}\right)}= 
    \left.\arg\min\left[\frac{1}{2}\left\Vert f\right\Vert _{K}^{2}\right]\right|_{x_*}=0
\end{gather}
since for Gaussian distributions it holds that the average case is also the most probable case. For the covariance we get
\begin{gather}
    \left\langle f(x)f(y) \right\rangle = \left.\frac{\delta^2\log(Z_0[\alpha])}{\delta\alpha\left(x\right)\delta\alpha\left(y\right)}\right|_{\alpha=0}=
    \frac{\int \D f\cdot f\left(x\right)\cdot f\left(y\right)\cdot\exp\left(-\frac{1}{2}\left\Vert f\right\Vert _{K}^{2}\right)}{\int \mathrm{D} f\exp\left(-\frac{1}{2}\left\Vert f\right\Vert _{K}^{2}\right)}=\\ \nonumber
    =\frac{\intop\prod_{i}df_{i}\cdot\sum_{i}f_{i}\phi_{i}\left(x\right)\cdot\sum_{j}f_{i}\phi_{j}\left(x\right)\cdot\exp\left(-\frac{1}{2}\sum_{l}\frac{f_{l}^{2}}{\lambda_{l}}\right)}{\intop\prod_{i}df_{i}\exp\left(-\frac{1}{2}\sum_{l}\frac{f_{l}^{2}}{\lambda_{l}}\right)}=\\ \nonumber
    =\sum_{i}\underbrace{\frac{\intop df\cdot f^{2}\cdot\exp\left(-\frac{f^{2}}{2\lambda_{i}}\right)}{\intop df\exp\left(-\frac{f^{2}}{2\lambda_{i}}\right)}}_{\lambda_{i}}\phi_{i}\left(x\right)\phi_{i}\left(y\right)+ \\ \nonumber
    \sum_{i\neq j}\underbrace{\frac{\intop df\cdot f\cdot\exp\left(-\frac{f^{2}}{2\lambda_{i}}\right)}{\intop df\exp\left(-\frac{f^{2}}{2\lambda_{i}}\right)}}_{0}\cdot\underbrace{\frac{\intop df\cdot f\cdot\exp\left(-\frac{f^{2}}{2\lambda_{j}}\right)}{\intop df\exp\left(-\frac{f^{2}}{2\lambda_{j}}\right)}}_{0}= \\ \nonumber =\sum_{i}\lambda_{i}\phi_{i}\left(x\right)\phi_{i}\left(y\right)=K\left(x,y\right)
\end{gather}

Indeed, $Z$ is the partition function corresponding to a centered GP with kernel $K$.

\subsection{With Data}
\label{Appendix:GprFieldTheoryPred}

We continue by establishing the exact equivalence between Bayesian inference on a GP and the corresponding partition function. 

From \ref{Eq:partitionNoData} we get that
\begin{equation}
    P[f]\propto\exp\left(-\frac{1}{2}\left\Vert f\right\Vert _{K}^{2}\right)
\end{equation}
For given target function $g$ and a sampled datapoint $\left(x_1,g(x_i)\right)$, assuming that $f$ is our prediction it holds that $g(x_i)\sim\mathcal{N}\left(f\left(x_i\right),\sigma^2\right)$, since $g$ distributes normally around $f$ with variance $\sigma^2$. Therefore, $p\left(g(x_i)|f\right)\propto\exp\left(-\left(g(x_i)-f\left(x_i\right)\right)^2/2\sigma^2\right)$, so 
\begin{equation}
\label{Eq:dataProb}
    P[D|f]=\prod_{i=1}^{N}p\left(g(x_i)|f,M\right)\propto\exp\left(-\frac{1}{2\sigma^2}\sum_{i=1}^{N}\left(g(x_i)-f\left(x_i\right)\right)^2\right)
\end{equation}
and using Bayes' theorem we get
\begin{equation}
    P[f|D]\propto\exp\left(-\frac{1}{2}\left\Vert f\right\Vert _{K}^{2}-\frac{1}{2\sigma^2}\sum_{i=1}^{N}\left(g(x_i)-f\left(x_i\right)\right)^2\right)
\end{equation}
which gives rise to the posterior partition function
\begin{equation}
    Z[\alpha]=\int\mathrm{D} f\exp\left({-\frac{1}{2}\left\Vert f\right\Vert_K^2} - \frac{1}{2\sigma^2}\sum_{i=1}^{N}\left(g(x_i)-f\left(x_i\right)\right)^2 +\int dx \alpha(x)f(x)\right) 
\end{equation}
and again, the exponent is quadratic in $f$ leading to a Gaussian distribution over the space of functions. Indeed, for the mean we get
\begin{gather}
\label{Eq:Zprediction}
    g^*(x^*)=\left\langle f(x_*) \right\rangle =\left.\frac{\delta\log(Z[\alpha])}{\delta\alpha\left(x_*\right)}\right|_{\alpha=0}=\\ \nonumber =\left.\arg\min\left[\frac{1}{2}\left\Vert f\right\Vert _{K}^{2}+\frac{1}{2\sigma^2}\sum_{i=1}^N\left(f(x_i)-g(x_i) \right)^2\right]\right|_{x_*}
\end{gather}
in agreement with \cite{Rasmussen2005}.

\subsection{Calculating Observables}

\subsubsection{Averaging $g^*$}

Applying the replica trick to Eq. \ref{Eq:Zprediction} and averaging over all the datasets of size $N$ we obtain
\begin{equation}
\label{Eq:Replica_f}
    \langle g^*(x_*)\rangle_{D_N}=\lim_{M\to 0} \frac{1}{M} \left.\frac{\delta \left\langle Z^M\left[\alpha\right]\right\rangle _{D_N}}{\delta\alpha\left(x_*\right)}\right|_{\alpha=0}
\end{equation}
for integer $M$ we get
\begin{gather}
\label{Eq:Z^M}
    Z^{M}\left[\alpha\right]=\underbrace{\int\dots\int}_{M}\prod_{j=1}^{M}\mathrm{D} f_{j} \\ \nonumber \exp\left(-\frac{1}{2}\sum_{j=1}^{M}\left\Vert f_{j}\right\Vert _{K}^{2}-\sum_{j=1}^{M}\sum_{i=1}^{N}\frac{\left(f_{j}\left(x_{i}\right)-g\left(x_{i}\right)\right)^{2}}{2\sigma^{2}}+\sum_{j=1}^{M}\int\alpha\left(x\right)f_{j}\left(x\right)dx\right)
\end{gather}
and after averaging
\begin{gather}
    \left\langle Z^M\left[\alpha\right]\right\rangle _{D_N} = \underbrace{\int...\int}_{M}\prod_{j=1}^{M}\mathrm{D} f_{j} \\ \nonumber \exp\left(-\frac{1}{2}\sum_{j=1}^{M}\left\Vert f_{j}\right\Vert _{K}^{2}+\sum_{j=1}^{M}\int\alpha\left(x\right)f_{j}\left(x\right)dx\right)\left\langle \exp\left(-\sum_{j=1}^{M}\frac{\left(f_{j}\left(x\right)-g\left(x\right)\right)^{2}}{2\sigma^{2}}\right)\right\rangle _{x\sim\mu}^{N}
\end{gather}
where $\langle\ldots\rangle_{x\sim\mu}=\int\ldots d\mu_x$.

Performing the Poissonic averging we get
\begin{gather}
\label{Eq:GCZM}
    \left\langle Z^M\left[\alpha\right]\right\rangle_\eta =e^{-\eta}\sum_{N=0}^\infty \frac{\eta^N}{N!} \left\langle Z^M\left[\alpha\right]\right\rangle _{D_N} = \\ \nonumber
    =\underbrace{\int\dots\int}_{M\,\,\mathrm{times}}\mathrm{D} f_{1}\dots\mathrm{D} f_{M} \\ \nonumber \exp\left(-\frac{1}{2}\sum_{j=1}^{M}\left\Vert f_{j}\right\Vert _{K}^{2}+\sum_{j=1}^{M}\int\alpha\left(x\right)f_{j}\left(x\right)dx+\eta\left\langle \exp\left(-\sum_{j=1}^{M}\frac{\left(f_{j}\left(x\right)-g\left(x\right)\right)^{2}}{2\sigma^{2}}\right)-1\right\rangle _{x\sim\mu}\right)
\end{gather} 
so overall
\begin{equation}
\label{Eq:fGC}
    \left\langle g^*\left(x_*\right) \right\rangle_\eta = \lim_{M\to0}\frac{1}{M}\left.\frac{\delta\langle Z^M[\alpha]\rangle_\eta}{\delta\alpha\left(x_*\right)}\right|_{\alpha=0}
\end{equation}
 
\subsubsection{Averaging ${g^*}^2$}

From \ref{Eq:Replica_f} we get that
\begin{equation}
\langle {g^*}^2(x_*)\rangle_{D_N}=\lim_{M\to 0} \lim_{W\to 0} \frac{1}{MW} \left.\frac{\delta^2 \langle Z^M \left[\alpha\right] Z^W \left[\beta\right]\rangle_{D_N}}{\delta\alpha\left(x_*\right)\delta\beta\left(x_*\right)}\right|_{\alpha,\beta=0}
\end{equation}
Therefore
\begin{equation}
\label{Eq:<<f^2>>}
\langle {g^*}^2(x_*)\rangle_\eta=\lim_{M\to 0} \lim_{W\to 0} \frac{1}{MW} \left.\frac{\delta^2 \langle Z^M \left[\alpha\right] Z^W \left[\beta\right]\rangle_\eta}{\delta\alpha\left(x_*\right)\delta\beta\left(x_*\right)}\right|_{\alpha,\beta=0}
\end{equation}
 
\section{Equivalence Kernel as Free Theory}
\label{Appendix:EKasFreeTheory}
Expending the nested exponent in Eq. \ref{Eq:GCZM} using (first order) Taylor series we get

\begin{gather}
    \left\langle Z^M\left[\alpha\right]\right\rangle _\eta=e^{-\eta}\underbrace{\int...\int}_{M}\prod_{j=1}^{M}\mathrm{D} f_{j} \\ \nonumber
    \exp\left(-\frac{1}{2}\sum_{j=1}^{M}\left\Vert f_{j}\right\Vert _{K}^{2}+\sum_{j=1}^{M}\int\alpha\left(x\right)f_{j}\left(x\right)dx+\eta\left\langle \exp\left(-\sum_{j=1}^{M}\frac{\left(f_{j}\left(x\right)-g\left(x\right)\right)^{2}}{2\sigma^{2}}\right)\right\rangle _{x\sim\mu}\right)= \\ \nonumber
    = \underbrace{\int...\int}_{M}\prod_{j=1}^{M}\mathrm{D} f_{j}
    \exp\left(-\frac{1}{2}\sum_{j=1}^{M}\left\Vert f_{j}\right\Vert _{K}^{2}+\sum_{j=1}^{M}\int\alpha\left(x\right)f_{j}\left(x\right)dx-\eta\left\langle \sum_{j=1}^{M}\frac{\left(f_{j}\left(x\right)-g\left(x\right)\right)^{2}}{2\sigma^{2}}\right\rangle _{x\sim\mu}\right) + O(1/\eta^2)= \\ \nonumber
    =\left[\int Df\exp\left(-\frac{1}{2}\left\Vert f\right\Vert _{K}^{2}+\int\alpha\left(x\right)f\left(x\right)d\mu_{x}-\frac{\eta}{2\sigma^{2}}\int d\mu_{x}\left(f\left(x\right)-g\left(x\right)\right)^{2}\right)\right]^{M} + O(1/\eta^2)= \\ \nonumber =\left( Z_{EK}\left[\alpha\right]\right)^{M} + O(1/\eta^2)
\end{gather}
where we defined
\begin{equation}
     Z_{EK}\left[\alpha\right] \defeq \int \D f\exp\left(-\frac{1}{2}\left\Vert f\right\Vert _{K}^{2}+\int\alpha\left(x\right)f\left(x\right)d\mu_{x}-\frac{\eta}{2\sigma^{2}}\int d\mu_{x}\left(f\left(x\right)-g\left(x\right)\right)^{2}\right)
\end{equation}

under this approximation we get
\begin{equation}
     \lim_{M\to0}\frac{\left\langle Z^M\left[\alpha\right]\right\rangle _\eta-1}{M}=
     \lim_{M\to0}\frac{\left( Z_{EK}\left[\alpha\right]\right)^{M}-1}{M} + O(1/\eta^2)=
     \log\left( Z_{EK}\left[\alpha\right]\right) + O(1/\eta^2) 
\end{equation}
Denoting the average w.r.t $ Z_{EK}$ as $\langle\dots\rangle_0$, The mean of the distribution induced by $ Z_{EK}$ is 
\begin{gather}
    \left\langle f\left(x_*\right)\right\rangle_0 = \left.\frac{\delta\log\left( Z_{EK}\left[\alpha\right]\right)}{\delta\alpha\left(x_{*}\right)}\right|_{\alpha=0}=\\ \nonumber
    =\frac{\int \D f\cdot f\left(x_{*}\right)\exp\left(-\frac{1}{2}\left\Vert f\right\Vert _{K}^{2}-\frac{\eta}{2\sigma^{2}}\int d\mu_{x}\left(f\left(x\right)-g\left(x\right)\right)^{2}\right)}{\int \D f\exp\left(-\frac{1}{2}\left\Vert f\right\Vert _{K}^{2}-\frac{\eta}{2\sigma^{2}}\int d\mu_{x}\left(f\left(x\right)-g\left(x\right)\right)^{2}\right)}= \\ \nonumber =\left.\arg\min\left[\frac{1}{2}\left\Vert f\right\Vert _{K}^{2}+\frac{\eta}{2\sigma^{2}}\int d\mu_{x}\left(f\left(x\right)-g\left(x\right)\right)^{2}\right]\right|_{x_*} = g^*_{EK,\eta}\left(x_*\right)
\end{gather}
where the last equality is due to \cite{Rasmussen2005}.

The covariance induced by $ Z_{EK}$ is
\begin{align}
\label{Eq:cov0}
    \mathrm{Cov}_0\left[f(x),f(y)\right]&=
    \left\langle f(x)f(y)\right\rangle_0 -\left\langle f(x)\right\rangle_0\left\langle f(y)\right\rangle_0 = \left.\frac{\delta^2\log\left( Z_{EK}\left[\alpha\right]\right)}{\delta\alpha(x)\delta\alpha(y)}\right|_{\alpha=0}\\ \nonumber
    &\stackrel{*}{=} \frac{\int Df\cdot f\left(x\right)f\left(y\right)\exp\left(-\frac{1}{2}\left\Vert f\right\Vert _{K}^{2}-\frac{\eta}{2\sigma^{2}}\int d\mu_{x}f^{2}\left(x\right)\right)}{\int Df\exp\left(-\frac{1}{2}\left\Vert f\right\Vert _{K}^{2}-\frac{\eta}{2\sigma^{2}}\int d\mu_{x}f^{2}\left(x\right)\right)} \\ \nonumber &\stackrel{**}{=}\frac{\intop\prod_{i}df_{i}\cdot\sum_{i,j}f_{i}f_{j}\phi_{i}\left(x\right)\phi_{j}\left(y\right)\cdot\exp\left(-\frac{1}{2}\sum_{i}\left(\frac{1}{\lambda_{i}}+\frac{\eta}{\sigma^{2}}\right)f_{i}^{2}\right)}{\intop\prod_{i}df_{i}\exp\left(-\frac{1}{2}\sum_{i}\left(\frac{1}{\lambda_{i}}+\frac{\eta}{\sigma^{2}}\right)f_{i}^{2}\right)} \\ \nonumber 
    &\stackrel{***}{=}\sum_{i}\frac{\intop df\cdot f^{2}\cdot\exp\left(-\frac{1}{2}\left(\frac{1}{\lambda_{i}}+\frac{\eta}{\sigma^{2}}\right)f^{2}\right)}{\intop df\exp\left(-\frac{1}{2}\left(\frac{1}{\lambda_{i}}+\frac{\eta}{\sigma^{2}}\right)f^{2}\right)}\phi_{i}\left(x\right)\phi_{i}\left(y\right) \\ \nonumber &=\sum_{i}\left(\frac{1}{\lambda_{i}}+\frac{\eta}{\sigma^{2}}\right)^{-1}\phi_{i}\left(x\right)\phi_{i}\left(y\right)
\end{align}
 where in ($*$) the non-centered part of the distribution was deleted, in ($**$) the eigenfunctions of $K$ were chosen as a base for the path integration and in ($***$) we used the fact that $\intop df\cdot f \cdot\exp\left(-\frac{1}{2}\left(\frac{1}{\lambda_{i}}+\frac{\eta}{\sigma^{2}}\right)f^{2}\right)=0$, since it is the mean of a centered (unnormalized) Gaussian distribution.
 
 For a rotationally invariant kernel, the eigenfunctions are $Y_{lm}$ and the eigenvalues are $\lambda_l$ (independent of $m$) so Eq. \ref{Eq:cov0} becomes
 \begin{gather}
     \mathrm{Cov}_0\left[f(x),f(y)\right]=\sum_{l}\left(\frac{1}{\lambda_{l}}+\frac{\eta}{\sigma^{2}}\right)^{-1}\underbrace{\sum_{m}Y_{lm}\left(x\right)Y_{lm}\left(y\right)}_{\deg(l)} \defeq C_{K,\sigma^2/\eta}
 \end{gather}
 which is a constant (independent of $x$ and $y$).

\clearpage
\section{Next Order Correction}
\label{App:NextOrderCorrections}
We now wish to perform the first order correction to the free theory. \\
Expanding Eq. \ref{Eq:GCZM} to the next order (keeping terms up to $O\left(1/\eta^2\right)$) we get
\begin{gather}
\label{Eq:pertZ}
    \left\langle  Z^{M}\left[\alpha\right]\right\rangle_\eta =\underbrace{\int...\int}_{M\,\,\mathrm{times}}\D f_{1}\dots\D f_{M}\\ \nonumber\exp\left(-\frac{1}{2}\sum_{j=1}^{M}\left\Vert f_{j}\right\Vert _{K}^{2}+\sum_{j=1}^{M}\int\alpha\left(x\right)f_{j}\left(x\right)dx+\eta\left\langle -\sum_{j=1}^{M}\frac{\left(f_{j}\left(x\right)-g\left(x\right)\right)^{2}}{2\sigma^{2}}\right\rangle _{x\sim\mu_{x}}\right)\\ \nonumber\exp\left(\frac{\eta}{2}\left\langle \left(\sum_{j=1}^{M}\frac{\left(f_{j}\left(x\right)-g\left(x\right)\right)^{2}}{2\sigma^{2}}\right)^{2}\right\rangle _{x\sim\mu_{x}}\right) + O\left( 1/\eta^3 \right) = \\ \nonumber=\underbrace{\int\dots\int}_{M\,\,\mathrm{times}}\D f_{1}\dots\D f_{M}\\ \nonumber\exp\left(\sum_{i=1}^{M}\left(-\frac{1}{2}\left\Vert f_{i}\right\Vert _{K}^{2}+\int\alpha\left(x\right)f_{i}\left(x\right)dx-\eta\left\langle \frac{\left(f_{i}\left(x\right)-g\left(x\right)\right)^{2}}{2\sigma^{2}}\right\rangle _{x\sim\mu}\right)\right)\\ \nonumber\exp\left(\frac{\eta}{8\sigma^{4}}\sum_{i=1}^{M}\sum_{j=1}^{M}\left(f_{j}\left(x\right)-g\left(x\right)\right)^{2}\cdot\left(f_{i}\left(x\right)-g\left(x\right)\right)^{2}\right) + O\left( 1/\eta^3 \right) = \\ \nonumber = \underbrace{\int\dots\int}_{M\,\,\mathrm{times}}\D f_{1}\dots\D f_{M}\\ \nonumber\exp\left(\sum_{i=1}^{M}\left(-\frac{1}{2}\left\Vert f_{i}\right\Vert _{K}^{2}+\int\alpha\left(x\right)f_{i}\left(x\right)dx-\eta\left\langle \frac{\left(f_{i}\left(x\right)-g\left(x\right)\right)^{2}}{2\sigma^{2}}\right\rangle _{x\sim\mu}\right)\right)\\ \nonumber\left(1+\frac{\eta}{8\sigma^{4}}\sum_{i=1}^{M}\sum_{j=1}^{M}\left(f_{j}\left(x\right)-g\left(x\right)\right)^{2}\cdot\left(f_{i}\left(x\right)-g\left(x\right)\right)^{2}\right) + O\left( 1/\eta^3 \right)
\end{gather}

\subsection{Calculating $\left\langle g^*\right\rangle_\eta$}

We now wish to calculate the correction to $\left\langle g^*\right\rangle_\eta$ given by Eq. \ref{Eq:pertZ}. From Eq. \ref{Eq:fGC} we get
\begin{gather}
\left\langle g^*\right\rangle_\eta =  g^*_{EK,\eta}(x_*)+ \\ \nonumber
\lim_{M\to0}\frac{1}{M}\frac{\eta}{8\sigma^{4}}\intop d\mu_{x}\left\langle \sum_{j=1}^{M}\sum_{l=1}^{M}\sum_{i=1}^{M}\left(f_{j}\left(x\right)-g\left(x\right)\right)^{2}\cdot\left(f_{l}\left(x\right)-g\left(x\right)\right)^{2}f_{i}\left(x_*\right)\right\rangle _{f_{1},\ldots,f_{M}\sim EK} + O\left( 1/\eta^3 \right)
\end{gather}

Simplifying the average of the multiple sums we get 
\begin{gather}
    \left\langle \sum_{j=1}^{M}\sum_{l=1}^{M}\sum_{i=1}^{M}\left(f_{j}\left(x\right)-g\left(x\right)\right)^{2}\cdot\left(f_{l}\left(x\right)-g\left(x\right)\right)^{2}f_{i}\left(x_*\right)\right\rangle _{f_{1},\ldots,f_{M}\sim EK} =\\ \nonumber
    =M\left\langle \left(f\left(x\right)-g\left(x\right)\right)^{4}f\left(x_*\right)\right\rangle _{0} \\ \nonumber+M\left(M-1\right)\left[2\left\langle \left(f\left(x\right)-g\left(x\right)\right)^{2}\right\rangle _{0}\left\langle \left(f\left(x\right)-g\left(x\right)\right)^{2}f\left(x_*\right)\right\rangle _{0}+\left\langle \left(f\left(x\right)-g\left(x\right)\right)^{4}\right\rangle _{0}\left\langle f\left(x_*\right)\right\rangle _{0}\right] \\ \nonumber
    +M\left(M-1\right)\left(M-2\right)\left\langle \left(f\left(x\right)-g\left(x\right)\right)^{2}\right\rangle _{0}^{2}\left\langle f\left(x_*\right)\right\rangle _{0}
\end{gather}

Since $f$ has a Gaussian distribution ($f\sim EK$), such averages can be calculated using Feynman diagrams. 

Let us denote $f(x)-g(x)$ by
\begin{tikzpicture}[x=0.75pt,y=0.75pt,yscale=-0.5,xscale=0.5]
\draw  [fill={rgb, 255:red, 0; green, 0; blue, 0 }  ,fill opacity=1 ] (309,41) -- (327.5,41) -- (327.5,59.5) -- (309,59.5) -- cycle ;
\end{tikzpicture} and $f(x_*)$ by
\begin{tikzpicture}[x=0.75pt,y=0.75pt,yscale=-0.5,xscale=0.5]
\draw  [color={rgb, 255:red, 0; green, 0; blue, 0 }  ,draw opacity=1 ][fill={rgb, 255:red, 0; green, 0; blue, 0 }  ,fill opacity=1 ] (368.02,49.98) .. controls (368.02,44.46) and (372.5,39.98) .. (378.02,39.98) .. controls (383.54,39.98) and (388.02,44.46) .. (388.02,49.98) .. controls (388.02,55.5) and (383.54,59.98) .. (378.02,59.98) .. controls (372.5,59.98) and (368.02,55.5) .. (368.02,49.98) -- cycle ;
\end{tikzpicture}
. Since our free theory is not centered ($\left\langle f \right\rangle_0= g^*_{EK,\eta}\neq 0$), we allow edges in the diagrams to be connected at only one side, representing the average of the vertex w.r.t the EK distribution. An edge connected to vertices on both sides represents the covariance. Note that since we divide by $M$ and take the limit $M\to0$, we do not care about diagrams which are not connected to $f(x_*)$ since they scale as $M^2$. 

Calculating the averages we get
\begin{gather}
\left\langle \left(f\left(x\right)-g\left(x\right)\right)^{4}f\left(x_*\right)\right\rangle _{0} = 
\end{gather}

\tikzset{every picture/.style={line width=0.75pt}} 

\begin{tikzpicture}[x=0.75pt,y=0.75pt,yscale=-1,xscale=1]

\draw  [color={rgb, 255:red, 0; green, 0; blue, 0 }  ,draw opacity=1 ][fill={rgb, 255:red, 0; green, 0; blue, 0 }  ,fill opacity=1 ] (193.02,29.98) .. controls (193.02,24.46) and (197.5,19.98) .. (203.02,19.98) .. controls (208.54,19.98) and (213.02,24.46) .. (213.02,29.98) .. controls (213.02,35.5) and (208.54,39.98) .. (203.02,39.98) .. controls (197.5,39.98) and (193.02,35.5) .. (193.02,29.98) -- cycle ;
\draw  [fill={rgb, 255:red, 0; green, 0; blue, 0 }  ,fill opacity=1 ] (134,21) -- (152.5,21) -- (152.5,39.5) -- (134,39.5) -- cycle ;
\draw   (133.88,31.18) .. controls (128.32,31.11) and (123.82,26.58) .. (123.82,21) .. controls (123.82,15.38) and (128.38,10.82) .. (134,10.82) .. controls (139.59,10.82) and (144.13,15.34) .. (144.18,20.92) -- (144.18,20.92) .. controls (144.13,15.34) and (139.59,10.82) .. (134,10.82) .. controls (128.38,10.82) and (123.82,15.38) .. (123.82,21) .. controls (123.82,26.58) and (128.32,31.11) .. (133.88,31.18) -- cycle ;
\draw    (143.25,30.25) -- (203.02,29.98) ;

\draw    (143.25,30.25) -- (143.25,51.25) ;

\draw  [color={rgb, 255:red, 0; green, 0; blue, 0 }  ,draw opacity=1 ][fill={rgb, 255:red, 0; green, 0; blue, 0 }  ,fill opacity=1 ] (348.02,29.98) .. controls (348.02,24.46) and (352.5,19.98) .. (358.02,19.98) .. controls (363.54,19.98) and (368.02,24.46) .. (368.02,29.98) .. controls (368.02,35.5) and (363.54,39.98) .. (358.02,39.98) .. controls (352.5,39.98) and (348.02,35.5) .. (348.02,29.98) -- cycle ;
\draw  [fill={rgb, 255:red, 0; green, 0; blue, 0 }  ,fill opacity=1 ] (289,21) -- (307.5,21) -- (307.5,39.5) -- (289,39.5) -- cycle ;
\draw    (277.5,30.31) -- (358.02,29.98) ;

\draw    (298.5,9.31) -- (298.25,50.25) ;

\draw (244,28.31) node   {$+$};
\draw (404,28.31) node   {$+$};
\draw (519,29.31) node  [align=left] {disconnected diagrams};
\draw (77,27.31) node   {$=$};
\draw (637,27.31) node   {$=$};

\end{tikzpicture}
\begin{gather*}=12\left( g^*_{EK,\eta}\left(x\right)-g\left(x\right)\right)\mathrm{Var}_0 \left[f\left(x\right)\right]\mathrm{Cov}_0 \left[f\left(x\right),f\left(x_{*}\right)\right] \\ \nonumber
+ 4\left( g^*_{EK,\eta}\left(x\right)-g\left(x\right)\right)^{3}\mathrm{Cov}_0 \left[f\left(x\right),f\left(x_{*}\right)\right] \\ \nonumber 
+ \text{ disconnected diagrams}
\end{gather*}
\begin{gather}
\left\langle \left(f\left(x\right)-g\left(x\right)\right)^{2}\right\rangle _{0}\left\langle \left(f\left(x\right)-g\left(x\right)\right)^{2}f\left(x_*\right)\right\rangle _{0} = 
\end{gather}
\begin{tikzpicture}[x=0.75pt,y=0.75pt,yscale=-1,xscale=1]

\draw  [fill={rgb, 255:red, 0; green, 0; blue, 0 }  ,fill opacity=1 ] (133,29) -- (151.5,29) -- (151.5,47.5) -- (133,47.5) -- cycle ;
\draw    (142.5,18.31) -- (142.25,59.25) ;

\draw  [fill={rgb, 255:red, 0; green, 0; blue, 0 }  ,fill opacity=1 ] (193,29) -- (211.5,29) -- (211.5,47.5) -- (193,47.5) -- cycle ;
\draw   (221.5,24.14) -- (221.5,51.57) .. controls (221.5,55.13) and (217.13,58.02) .. (211.75,58.02) .. controls (206.37,58.02) and (202,55.13) .. (202,51.57) -- (202,24.14) .. controls (202,20.58) and (206.37,17.69) .. (211.75,17.69) .. controls (217.13,17.69) and (221.5,20.58) .. (221.5,24.14) -- cycle ;
\draw  [color={rgb, 255:red, 0; green, 0; blue, 0 }  ,draw opacity=1 ][fill={rgb, 255:red, 0; green, 0; blue, 0 }  ,fill opacity=1 ] (351.02,37.98) .. controls (351.02,32.46) and (355.5,27.98) .. (361.02,27.98) .. controls (366.54,27.98) and (371.02,32.46) .. (371.02,37.98) .. controls (371.02,43.5) and (366.54,47.98) .. (361.02,47.98) .. controls (355.5,47.98) and (351.02,43.5) .. (351.02,37.98) -- cycle ;
\draw  [fill={rgb, 255:red, 0; green, 0; blue, 0 }  ,fill opacity=1 ] (292,29) -- (310.5,29) -- (310.5,47.5) -- (292,47.5) -- cycle ;
\draw    (280.5,38.31) -- (361.02,37.98) ;

\draw (172,35.31) node   {$+$};
\draw (118,37.02) node [scale=2.488]  {$($};
\draw (243,37.02) node [scale=2.488]  {$)$};
\draw (265,37.02) node [scale=2.488]  {$($};
\draw (596,37.02) node [scale=2.488]  {$)$};
\draw (403,36.31) node   {$+$};
\draw (506,38.31) node  [align=left] {disconnected diagrams};
\draw (637,37.31) node   {$=$};
\draw (77,37.31) node   {$=$};

\end{tikzpicture}
\begin{gather*}
=2\mathrm{Cov}_0 \left[f\left(x\right),f\left(x_{*}\right)\right]\left( g^*_{EK,\eta}\left(x\right)-g\left(x\right)\right)^{3} \\ \nonumber
+2\mathrm{Var}_0 \left[f\left(x\right)\right]\mathrm{Cov}_0 \left[f\left(x\right),f\left(x_{*}\right)\right]\left( g^*_{EK,\eta}\left(x\right)-g\left(x\right)\right) \\ \nonumber
+ \text{ disconnected diagrams}
\end{gather*}
\begin{gather}
\left\langle \left(f\left(x\right)-g\left(x\right)\right)^{4}\right\rangle _{0}\left\langle f\left(x_*\right)\right\rangle _{0} = \text{disconnected diagrams}
\end{gather}
\begin{gather}
\left\langle \left(f\left(x\right)-g\left(x\right)\right)^{2}\right\rangle _{0}^{2}\left\langle f\left(x_*\right)\right\rangle _{0}= \text{disconnected diagrams}
\end{gather}
Taking the limit $M\to0$ and summing everything together we get
\begin{gather}
\label{Eq:Relevant2f2}
\lim_{M\to0}\frac{1}{M}\left\langle \sum_{j=1}^{M}\sum_{l=1}^{M}\sum_{i=1}^{M}\left(f_{j}\left(x\right)-g\left(x\right)\right)^{2}\cdot\left(f_{l}\left(x\right)-g\left(x\right)\right)^{2}f_{i}\left(x_*\right)\right\rangle _{f_{1},\ldots,f_{M}\sim EK}
    =\\ \nonumber
    8\left( g^*_{EK,\eta}\left(x\right)-g\left(x\right)\right)\mathrm{Var}_{0}\left[f\left(x\right)\right]\mathrm{Cov}_{0}\left[f\left(x\right),f\left(x_*\right)\right]
\end{gather}
so finally

\begin{gather}
\left\langle g^*\right\rangle_\eta = \\ \nonumber g^*_{EK,\eta}\left(x_{*}\right)+\frac{\eta}{\sigma^{4}}\intop d\mu_{x}\left( g^*_{EK,\eta}\left(x\right)-g\left(x\right)\right)\mathrm{Var}_{0}\left[f\left(x\right)\right]\mathrm{Cov}_{0}\left[f\left(x\right),f\left(x_*\right)\right] + O\left( 1/\eta^3 \right)
\end{gather}

Substituting the expressions for the free variance and the covariance (Eq. \ref{Eq:cov0}) we get

\begin{gather}
\label{Eq:f_nextOrder}
    \left\langle g^*\right\rangle_\eta = \\ \nonumber
     g^*_{EK,\eta}\left(x_*\right)-\frac{\eta}{\sigma^{4}}\sum_{i,j,k}\frac{\frac{\sigma^{2}}{\eta}}{\lambda_{i}+\frac{\sigma^{2}}{\eta}}\left(\frac{1}{\lambda_{j}}+\frac{\eta}{\sigma^{2}}\right)^{-1}\left(\frac{1}{\lambda_{k}}+\frac{\eta}{\sigma^{2}}\right)^{-1}g_{i}\phi_{j}\left(x_*\right)\intop d\mu_{x}\phi_{i}\left(x\right)\phi_{j}\left(x\right)\phi_{k}^{2}\left(x\right) + O\left( 1/\eta^3 \right)
\end{gather}

For a rotationally invariant kernel, \ref{Eq:f_nextOrder} becomes

\begin{equation}
\label{Eq:f_nextOrder_rot}
\left\langle g^*\right\rangle_\eta =
 g^*_{EK,\eta}\left(x_*\right)-\sum_{l,m}\frac{\eta^{-1} \lambda_l C_{K,\sigma^2/\eta}}{(\lambda_l+\sigma^2/\eta)^2}g_{lm}Y_{lm}\left(x_*\right) + O\left( 1/\eta^3 \right)
\end{equation}

\subsection{Calculating $\left\langle {g^*}^2\right\rangle_\eta$}
Substituting \ref{Eq:Z^M} in \ref{Eq:<<f^2>>} we get
\begin{gather}
\left\langle {g^*}^2\right\rangle_\eta = 
\lim_{M\to0}\lim_{W\to0}\frac{1}{MW\cdot\left(Z_{EK}\left[\alpha=0\right]\right)^{M+W}}\cdot\int Df_{1}\ldots\int Df_{M}\int D\tilde{f}_{1}\ldots\int D\tilde{f}_{W} \\ \nonumber \exp\left(-\eta-\frac{1}{2}\sum_{m=1}^{M}\left\Vert f_{m}\right\Vert _{K}^{2}-\frac{1}{2}\sum_{w=1}^{W}\left\Vert \tilde{f}_{w}\right\Vert _{K}^{2}\right) \\ \nonumber \exp\left(\eta\left\langle \exp\left(-\sum_{m=1}^{M}\frac{\left(f_{m}\left(x\right)-g\left(x\right)\right)^{2}}{2\sigma^{2}}-\sum_{w=1}^{W}\frac{\left(\tilde{f}_{w}\left(x\right)-g\left(x\right)\right)^{2}}{2\sigma^{2}}\right)\right\rangle _{x\sim\mu_{x}}\right) \sum_{m=1}^{M}f_{m}\left(x_{*}\right)\sum_{w=1}^{W}\tilde{f}_{w}\left(x_{*}\right)
\end{gather}
By expanding to the same order we get (all equalities are up to $O\left( 1/\eta^3 \right)$)
\begin{gather}
\left\langle {g^*}^2\left(x_{*}\right) \right\rangle_\eta    =\lim_{M\to0}\lim_{W\to0}\frac{1}{MW\cdot\left(Z_{EK}\left[\alpha=0\right]\right)^{M+W}}\cdot\int Df_{1}\ldots\int Df_{M}\int D\tilde{f}_{1}\ldots\int D\tilde{f}_{W} \\ \nonumber \exp\left(-\frac{1}{2}\sum_{m=1}^{M}\left\Vert f_{m}\right\Vert _{K}^{2}-\frac{1}{2}\sum_{w=1}^{W}\left\Vert \tilde{f}_{w}\right\Vert _{K}^{2}+\eta\left\langle \left(-\sum_{m=1}^{M}\frac{\left(f_{m}\left(x\right)-g\left(x\right)\right)^{2}}{2\sigma^{2}}-\sum_{w=1}^{W}\frac{\left(\tilde{f}_{w}\left(x\right)-g\left(x\right)\right)^{2}}{2\sigma^{2}}\right)\right\rangle _{x\sim\mu_{x}}\right) \\ \nonumber \left(1+\frac{\eta}{2}\left\langle \left(\sum_{m=1}^{M}\frac{\left(f_{m}\left(x\right)-g\left(x\right)\right)^{2}}{2\sigma^{2}}+\sum_{w=1}^{W}\frac{\left(\tilde{f}_{w}\left(x\right)-g\left(x\right)\right)^{2}}{2\sigma^{2}}\right)^{2}\right\rangle _{x\sim\mu_{x}}\right)\sum_{m=1}^{M}f_{m}\left(x_{*}\right)\sum_{w=1}^{W}\tilde{f}_{w}\left(x_{*}\right) = \\ \nonumber =  {g^*}_{EK,\eta}^2\left(x_{*}\right)+ \\ \nonumber \lim_{M\to0}\lim_{W\to0}\frac{1}{MW}\cdot\frac{\eta}{8\sigma^{4}}\int d\mu_{x}\left\langle \left(\sum_{a=1}^{M}\left(f_{a}\left(x\right)-g\left(x\right)\right)^{2}+\sum_{b=1}^{W}\left(\tilde{f}_{b}\left(x\right)-g\left(x\right)\right)^{2}\right)^{2}\sum_{c=1}^{M}f_{c}\left(x_{*}\right)\sum_{d=1}^{W}\tilde{f}_{d}\left(x_{*}\right)\right\rangle_0 
  \\ \nonumber 
 = {g^*}_{EK,\eta}^{2}\left(x_{*}\right) +  \lim_{M\to0}\lim_{W\to0}\frac{1}{MW}\cdot\frac{\eta}{4\sigma^{4}}\int d\mu_{x} \\ \nonumber \left[\left\langle \sum_{a=1}^{M}\left(f_{a}\left(x\right)-g\left(x\right)\right)^{2}\sum_{b=1}^{M}\left(f_{b}\left(x\right)-g\left(x\right)\right)^{2}\sum_{c=1}^{M}f_{c}\left(x_{*}\right)\sum_{d=1}^{W}\tilde{f}_{d}\left(x_{*}\right)\right\rangle_0\right. \\ \nonumber +\left.\left\langle \sum_{a=1}^{M}\left(f_{a}\left(x\right)-g\left(x\right)\right)^{2}\sum_{b=1}^{W}\left(\tilde{f}_{b}\left(x\right)-g\left(x\right)\right)^{2}\sum_{c=1}^{M}f_{c}\left(x_{*}\right)\sum_{d=1}^{W}\tilde{f}_{d}\left(x_{*}\right)\right\rangle_0 \right] = \\ \nonumber = {g^*}_{EK,\eta}^{2}\left(x_{*}\right)+ \\ \nonumber \frac{\eta}{4\sigma^{4}}\int d\mu_{x}\lim_{M\to0}\frac{1}{M}\left\langle \sum_{a=1}^{M}\left(f_{a}\left(x\right)-g\left(x\right)\right)^{2}\sum_{b=1}^{M}\left(f_{b}\left(x\right)-g\left(x\right)\right)^{2}\sum_{c=1}^{M}f_{c}\left(x_{*}\right)\right\rangle_0  {g^*}_{EK,\eta}\left(x_{*}\right) \\ \nonumber +\frac{\eta}{4\sigma^{4}}\int d\mu_{x}\left(\lim_{M\to0}\frac{1}{M}\left\langle \sum_{a=1}^{M}\left(f_{a}\left(x\right)-g\left(x\right)\right)^{2}\sum_{b=1}^{M}f_{b}\left(x_{*}\right)\right\rangle_0 \right)^{2} 
\end{gather}
The first integrand was already calculated and is given in \ref{Eq:Relevant2f2}. For the second integrand we get
\begin{gather}
    \lim_{M\to0}\frac{1}{M}\left\langle \sum_{a=1}^{M}\left(f_{a}\left(x\right)-g\left(x\right)\right)^{2}\sum_{b=1}^{M}f_{b}\left(x_*\right)\right\rangle = \\ \nonumber
    =\lim_{M\to0}\frac{1}{M}\left[M\left\langle \left(f\left(x\right)-g\left(x\right)\right)^{2}f\left(x_*\right)\right\rangle +M\left(M-1\right)\left\langle f\left(x_*\right)\right\rangle \left\langle \left(f\left(x\right)-g\left(x\right)\right)^{2}\right\rangle \right]= \\ \nonumber
    \left\langle \left(f\left(x\right)-g\left(x\right)\right)^{2}f\left(x_*\right)\right\rangle -\left\langle f\left(x_*\right)\right\rangle \left\langle \left(f\left(x\right)-g\left(x\right)\right)^{2}\right\rangle =
\end{gather}
\centering
\begin{tikzpicture}[x=0.75pt,y=0.75pt,yscale=-1,xscale=1]

\draw  [color={rgb, 255:red, 0; green, 0; blue, 0 }  ,draw opacity=1 ][fill={rgb, 255:red, 0; green, 0; blue, 0 }  ,fill opacity=1 ] (240.02,40.98) .. controls (240.02,35.46) and (244.5,30.98) .. (250.02,30.98) .. controls (255.54,30.98) and (260.02,35.46) .. (260.02,40.98) .. controls (260.02,46.5) and (255.54,50.98) .. (250.02,50.98) .. controls (244.5,50.98) and (240.02,46.5) .. (240.02,40.98) -- cycle ;
\draw  [fill={rgb, 255:red, 0; green, 0; blue, 0 }  ,fill opacity=1 ] (181,32) -- (199.5,32) -- (199.5,50.5) -- (181,50.5) -- cycle ;
\draw    (169.5,41.31) -- (250.02,40.98) ;

\draw (292,41.31) node   {$+$};
\draw (395,41.31) node  [align=left] {disconnected diagrams};
\draw (492,41.31) node   {$=$};
\draw (147,41.31) node   {$=$};

\end{tikzpicture}
\begin{gather*} 
    =2\left({g^*}_{EK,\eta}\left(x\right)-g\left(x\right)\right)\mathrm{Cov}_0\left[f\left(x\right),f\left(x_*\right)\right] = O\left( 1/\eta^3 \right)
\end{gather*}
\flushleft
so the correction for $\left\langle {g^*}^2\left(x_{*}\right) \right\rangle_\eta$ is
\begin{gather}
\label{Eq:f2_nextOrder}
    \left\langle {g^*}^2\left(x_{*}\right) \right\rangle_\eta = \\ \nonumber
    {g^*}^2_{EK,\eta}\left(x_*\right)-\underbrace{2\frac{\eta}{\sigma^{4}}g^*_{EK,\eta}\left(x_*\right)\sum_{i,j,k}\frac{\frac{\sigma^{2}}{\eta}}{\lambda_{i}+\frac{\sigma^{2}}{\eta}}\left(\frac{1}{\lambda_{j}}+\frac{\eta}{\sigma^{2}}\right)^{-1}\left(\frac{1}{\lambda_{k}}+\frac{\eta}{\sigma^{2}}\right)^{-1}g_{i}\phi_{j}\left(x_*\right)\intop d\mu_{x}\phi_{i}\left(x\right)\phi_{j}\left(x\right)\phi_{k}^{2}\left(x\right)}_{O\left( 1/\eta^2 \right)} + O\left( 1/\eta^3 \right)
\end{gather}
and for a rotationally invariant kernel we get
\begin{equation}
\label{Eq:f2_nextOrder_rot}
\left\langle {g^*}^2\left(x_{*}\right) \right\rangle_\eta =
{g^*}^2_{EK,\eta}\left(x_*\right)-2g^*_{EK,\eta}\left(x_*\right)\sum_{l,m}\frac{\eta^{-1} \lambda_l C_{K,\sigma^2/\eta}}{(\lambda_l+\sigma^2/\eta)^2}g_{lm}Y_{lm}\left(x_*\right) + O\left( 1/\eta^3 \right)
\end{equation}

\section{Various insights}
\label{Appendix:Insights}

\subsection{Correction means worse generalization}

The correction always means worse generalization than what the EK suggests. Indeed, expending equation \eqref{Eq:f_nextOrder_rot} we get

\begin{gather*}
\left\langle g^*\right\rangle_\eta =
{g^*}_{EK,\eta}\left(x_*\right)-\sum_{l,m}\frac{\eta^{-1} \lambda_l C_{K,\sigma^2/\eta}}{(\lambda_l+\sigma^2/\eta)^2}g_{lm}Y_{lm}\left(x_*\right) + O\left( 1/\eta^3 \right) = \\ 
=\sum_{l,m}\frac{\lambda_{l}}{\lambda_{l}+\frac{\sigma^{2}}{\eta}}g_{l,m}Y_{l,m}\left(x_*\right)-\sum_{l,m}\frac{\eta^{-1} \lambda_l C_{K,\sigma^2/\eta}}{(\lambda_l+\sigma^2/\eta)^2}g_{lm}Y_{lm}\left(x_*\right) + O\left( 1/\eta^3 \right) = \\
=\sum_{l,m}\underbrace{\left(\frac{\lambda_{l}}{\lambda_{l}+\frac{\sigma^{2}}{\eta}}-\underbrace{\frac{\eta^{-1} \lambda_l C_{K,\sigma^2/\eta}}{(\lambda_l+\sigma^2/\eta)^2}}_{\text{positive}}\right)}_{<\frac{\lambda_{l}}{\lambda_{l}+\frac{\sigma^{2}}{\eta}}<1}g_{l,m}Y_{l,m}\left(x_*\right)
\end{gather*}

\subsection{Exact eigenvalues for 2-layer ReLU NNGP and NTK with $\sigma^{2}_b=0$}

For the NNGP associated with a 2-layer ReLU NTK without bias we were able to fined an exact expression for the eigenvalues for all $l$:

\begin{gather*}
\lambda_{l=2k}=\sigma_{w_{0}}^{2}\sigma_{w_{1}}^{2}\cdot\frac{d}{16\pi^2}\left(\frac{\Gamma\left(\frac{l-1}{2}\right)\Gamma\left(\frac{d}{2}\right)}{\Gamma\left(\frac{l+d+1}{2}\right)}\right)^{2} \\ \nonumber
\lambda_{l=2k+1}=\sigma_{w_{0}}^{2}\sigma_{w_{1}}^{2}\cdot\frac{1}{4d}\delta_{k,0}
\end{gather*}

and for NTK:

\begin{gather*}
\label{Eq:EigsNTK}
\lambda_{2k}=\frac{\sigma_{w_{1}}^{2}\sigma_{w_{2}}^{2}}{2\pi}\cdot\frac{d(1+2k)+(1-2k)^{2}}{8 \pi}\left(\frac{\Gamma\left(k-\frac{1}{2}\right)\Gamma\left(\frac{d}{2}\right)}{\Gamma\left(k+\frac{d+1}{2}\right)}\right)^{2} \\ \nonumber
\lambda_{2k+1}=\frac{\sigma_{w_{1}}^{2}\sigma_{w_{2}}^{2}}{2\pi}\cdot\frac{\pi}{d}\delta_{k,0}
\end{gather*}

It is interesting to note that for all odd $l>1$ $\lambda_l=0$ so the expressive power of the kernel (and hence the neural network) is greatly reduced.








\clearpage
\section{Accuracy of the renormalized NTK}
\label{AppRG}
Let us consider the random variable $t=x\cdot x'$, for two normalized datapoints $x$ and $x'$ drawn uniformly from the unit hypersphere $S^{d-1}$. Without loss of generality, $x$ can be assumed to be a unit vector in the direction of the last axis, and therefore $t$ is the last component of $x'$. The density at $t\in[-1,1]$ is therefore proportional to the surface area lying at a height between $t$ and $t+dt$ on the unit sphere. That proportion occurs within a belt of height $dt$ and radius $\sqrt{1-t^2}$, which is a conical frustum constructed out of a $d-2$ dimensional hypersphere of radius $\sqrt{1-t^2}$, of height $dt$, and slope $1/\sqrt{1-t^2}$. Hence the probability is proportional to $p(t)dt\propto (1-t^2)^{(d-3)/2}dt$. Defining $u=(t+1)/2$ it holds that $p(u)du\propto u^{(d-3)/2}(1-u)^{(d-3)/2}$, meaning that $u\sim \mathrm{Beta} \left(\frac{d-1}{2},\frac{d-1}{2}\right)$, and for large $d$ it holds that $\mathrm{Var}[t]=O(d^{-1})$. Since $t$ is bounded to $[-1,1]$, the random variable $t^r$ must have a standard deviation which is a decaying function of $r$. Indeed, for $n \ll d$ and large $d$, approximating the integral $\int t^n (1-t^2)^{(d-3)/2}dt$ using saddle point approximation we get that $f(t) = n\ln(t)+\frac{d-3}{2}\ln(1-t^2)$ is maximal for $t_0=[n/(n+d-3)]^{1/2}$, and $f''(t_0)=2[n^2-(d-3)^2]/(d-3)$  so overall

\begin{equation}
\left\langle t^{2r}\right\rangle = 
\frac{\int t^{2r} (1-t^2)^{(d-3)/2}}
{\int t^0 (1-t^2)^{(d-3)/2}}
\approx \left[1-\left(\frac{2r}{d-3}\right)^2\right]^{-1/2} \left(\frac{2r}{2r+d-3}\right)^r \left(\frac{d-3}{2r+d-3}\right)^{(d-3)/2}
\end{equation}

This implies that for $r\ll d$, the standard deviation of $t^r$ is $O(d^{-r/2})$.
Considering next the tail of the Taylor expansion $\sum_{q > r} b_q (x \cdot x')^q$, projected on the dataset ($\sum_{q > r} b_q (x_n \cdot x_m)^q$). The resulting $N$ by $N$ matrix is $\sum_{q > r} b_q$ on the diagonal but $O(d^{-(r+1)/2})$ in all other entries. As we justified in the main text, our renormalization transformation amounts to keeping only the diagonal piece of this matrix and interpreting it as noise. 

Consider then \eqref{Eq:GPPred} for $g^*$ in two scenarios: (I) $g^*_{\infty}$ with the full NTK ($K(x,x')$) and no noise and (II) $g^{*}_{r}$ with the NTK trimmed after the $r$'th power ($K_r(x,x')$) but with $\sigma_r^2=\sum_{q > r}b_q$. The first $K(x_*,x_n)$ piece, for $x_*$ drawn from the dataset distribution, obeys $K(x_*,x_n)-K_r(x_*,x_n) = O(d^{-(r+1)/2})$. Next we compare $K_r(x_n,x_m) + I_{nm} \sigma_{r}^2$ and $K(x_n,x_n)$. On their diagonal they agree exactly but their off-diagonal terms agree only up to a $O(d^{-(r+1)/2})$ discrepancy. Denoting by $\delta K$ the difference between these two matrices, we may expand $K^{-1} = [K_r+\sigma_m^2{\rm I} + \delta K]^{-1}=[K_r+\sigma_r^2{\rm I}]^{-1}[1- \delta K[K_r+\sigma_r^2{\rm I}]^{-1}+\delta K[K_r+\sigma_r^2{\rm I}]^{-1}\delta K[K_r+\sigma_r^2{\rm I}]^{-1}+...]$. 

We next argue that $\delta K[K_r+\sigma_r^2{\rm I}]^{-1}$ multiplied by target vector ($g(x_n)$) is negligible compared to the identity for large enough $r$ thereby establishing the equivalence of the two scenarios. Indeed consider the eigenvalues of $\delta K[K_r+\sigma_r^2{\rm I}]^{-1}$. As $\delta K_{nm}$ is $O(d^{-(r+1)/2})$ its typical eigenvalues are $O(\sqrt{N} d^{-(r+1)/2})$ and bounded by $O(N d^{-(r+1)/2})$. The typical eigenvalues of $[K_r+\sigma_m^2{\rm I}]^{-1}$ are of the same order as $K(x_n,x_n)=K$ and bounded from below by $\sigma_r^2$. Thus typical eigenvalues of $\delta K[K_r+\sigma_r^2{\rm I}]^{-1}$ are $O(\sqrt{N} d^{-(r+1)/2}/K)$ and bounded from above by $O(N d^{-(r+1)/2}/\sigma_r^2)$. The NTK has the desirable property that $\sigma_r^2$ decays very slowly. Thus certainly in the typical case but even in the worse case scenario we expect good agreement at large $r$. In Fig. 1, right panel, we provide supporting numerical evidence. 

We refer to $K_r(x,x')$ as the renormalized NTKs at the scale $r$. As follows from \eqref{Eq:generalEigen}, $\lambda_l$'s with $l \geq r$ are zero. Therefore, as advertised, the high-energy-sector has been removed and compensated by noise on the target and a change of the remaining $l<r$ (low-energy) eigenvalues. A proper choice of $r$ involve two considerations. Requiring perturbation theory to hold well ($C_{K_r,\sigma_r^2/\eta} < \sigma_r^2$) which puts an $\eta$-depended upper bound on $r$ and requiring small discrepancy in predictions puts another $\eta$ dependent lower bound on $r$ (typically $\sqrt{N}d^{-(r+1)/2} \ll 1$). 

Lastly we comment that our renormalization NTK approach is not limited to uniform datasets. The entire logic relies on having a rapidly decaying ratio of off-diagonal moments ($(x_n \cdot x_m)^{2r}$) and diagonal moments $(x_n \cdot x_n)^{2r}$ as one increases $r$. We expect this to hold in real-world distributions. For instance for a multi-dimension Gaussian data distribution the input dimension ($d$) traded by an effective dimension ($d_{eff}$) defined by the variance of $(x_m \cdot x_n)$.  

\clearpage
\section{Hyper-parameter optimization experiment results}
\label{Appendix:hyperparameterExperimentResults}
\begin{table}[ht]
\centering
\caption{Hyper-parameter performance comparison}
\label{Tbl:hyperparameterFullTable}
\begin{ruledtabular}
\begin{tabular}{lrrrrrrrrr}
           &  Test & Prediction &   GPR & $\sigma_{r}^2$ &    Train & $\sigma_{w_1}$ & $\sigma_{b_1}$ & $\sigma_{w_2}$ & $\sigma_{b_2}$ \\
 Random  1 & 0.389 &      0.400 & 0.364 &        0.068 & 1.29e-04 &        1.555 &        0.032 &        2.028 &        0.026 \\
 Random  2 & 0.316 &      0.287 & 0.331 &        0.004 & 4.17e-04 &        0.914 &        0.029 &        0.880 &        0.062 \\
 Random  3 & 0.191 &      0.219 & 0.250 &        0.003 & 3.36e-04 &        0.922 &        0.049 &        0.760 &        0.046 \\
 Random  4 & 0.300 &      0.324 & 0.306 &        0.045 & 1.39e-04 &        1.552 &        0.058 &        1.644 &        0.054 \\
 Random  5 & 0.268 &      0.338 & 0.273 &        0.070 & 1.24e-04 &        1.585 &        0.072 &        2.020 &        0.029 \\
 Random  6 & 0.413 &      0.406 & 0.382 &        0.065 & 1.28e-04 &        2.037 &        0.032 &        1.512 &        0.053 \\
 Random  7 & 0.332 &      0.297 & 0.335 &        0.010 & 8.76e-04 &        0.994 &        0.030 &        1.190 &        0.071 \\
 Random  8 & 0.228 &      0.245 & 0.262 &        0.015 & 6.41e-04 &        1.165 &        0.059 &        1.271 &        0.068 \\
 Random  9 & 0.337 &      0.308 & 0.332 &        0.018 & 1.37e-03 &        0.909 &        0.027 &        1.758 &        0.030 \\
 Random 10 & 0.371 &      0.392 & 0.355 &        0.069 & 1.19e-04 &        1.658 &        0.041 &        1.908 &        0.057 \\
 Random 11 & 0.313 &      0.316 & 0.319 &        0.032 & 1.31e-04 &        1.440 &        0.050 &        1.502 &        0.045 \\
 Random 12 & 0.335 &      0.340 & 0.329 &        0.042 & 1.33e-04 &        2.106 &        0.065 &        1.175 &        0.040 \\
 Random 13 & 0.397 &      0.336 & 0.373 &        0.018 & 1.71e-03 &        1.546 &        0.029 &        1.037 &        0.044 \\
 Random 14 & 0.236 &      0.253 & 0.271 &        0.017 & 4.18e-04 &        1.388 &        0.067 &        1.133 &        0.050 \\
 Random 15 & 0.293 &      0.288 & 0.306 &        0.018 & 6.37e-04 &        1.534 &        0.058 &        1.068 &        0.065 \\
 Random 16 & 0.175 &      0.198 & 0.214 &        0.011 & 1.04e-03 &        1.132 &        0.074 &        1.117 &        0.053 \\
 Random 17 & 0.206 &      0.226 & 0.246 &        0.013 & 7.66e-04 &        1.095 &        0.061 &        1.275 &        0.072 \\
 Random 18 & 0.264 &      0.264 & 0.292 &        0.011 & 1.00e-03 &        1.035 &        0.043 &        1.230 &        0.066 \\
 Random 19 & 0.239 &      0.248 & 0.282 &        0.006 & 5.34e-04 &        0.835 &        0.037 &        1.107 &        0.050 \\
 Random 20 & 0.385 &      0.378 & 0.364 &        0.054 & 1.43e-04 &        1.810 &        0.039 &        1.540 &        0.046 \\
 Random 21 & 0.318 &      0.300 & 0.323 &        0.018 & 9.58e-04 &        1.282 &        0.042 &        1.266 &        0.055 \\
 Worst     & 0.413 &      0.406 & 0.382 &        0.065 & 1.28e-04 &        2.037 &        0.032 &        1.512 &        0.053 \\
 Median    & 0.313 &      0.316 & 0.319 &        0.032 & 1.31e-04 &        1.440 &        0.050 &        1.502 &        0.045 \\
 Best      & 0.175 &      0.198 & 0.214 &        0.011 & 1.04e-03 &        1.132 &        0.074 &        1.117 &        0.053 \\
 Typical   & 0.307 &      0.307 & 0.317 &        0.028 & 1.41e-04 &        1.414 &        0.050 &        1.414 &        0.050 \\
 Optimized & 0.078 &      0.110 & 0.141 &        0.002 & 1.66e-04 &        0.707 &        0.075 &        0.707 &        0.027 \\
\end{tabular}
\end{ruledtabular}
\end{table}

\begin{figure*}[h]
     \centering
     \includegraphics[width=\linewidth]{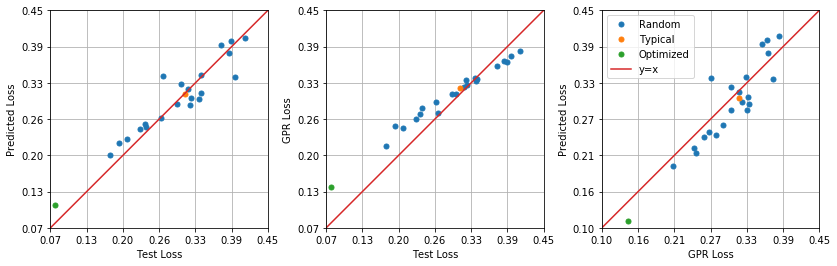}
     \label{Fig:hyperparameter_corr}
\end{figure*}

\end{document}